%% file: main.tex
\documentclass[sigconf]{acmart}
\AtBeginDocument{%
  \providecommand\BibTeX{{%
    \normalfont B\kern-0.5em{\scshape i\kern-0.25em b}\kern-0.8em\TeX}}}

\copyrightyear{2024}
\acmYear{2024}
\setcopyright{acmlicensed}
\acmDOI{10.1145/3637528.3671951}

\acmConference[KDD '24]{Proceedings of the 30th ACM SIGKDD Conference on Knowledge Discovery and Data Mining}{August 25--29, 2024}{Barcelona, Spain}

\acmBooktitle{Proceedings of the 30th ACM SIGKDD Conference on Knowledge Discovery and Data Mining (KDD '24), August 25--29, 2024, Barcelona, Spain}
\acmISBN{979-8-4007-0490-1/24/08}

\usepackage{enumitem}

\usepackage{amsthm}

\theoremstyle{definition}
\newtheorem{proposition}{Proposition}

\usepackage{float}

\usepackage{bbm}

\usepackage{mathtools}

\usepackage{multirow}

\usepackage{siunitx}

\usepackage{comment}

\usepackage{bm}

\usepackage{adjustbox}

\newcommand{\rowfonttype}{}% Current row font
\newcolumntype{L}{>{\rowfonttype\strut}l}
\newcolumntype{C}{>{\rowfonttype\strut}c}
\newcolumntype{R}{>{\rowfonttype\strut}r}

\usepackage{algorithm}
\usepackage{algorithmic}
\algsetup{
  linenodelimiter = {}
}

\newcommand{\algname}{\textit{CURLS}\xspace}

\usepackage{xspace,xpunctuate}

\newcommand{\ie}{\textit{i.e.},\xspace}
\newcommand{\etal}{\xspace\textit{et al.}\xspace}
\newcommand{\eg}{\textit{e.g.},\xspace}
\newcommand{\etc}{\textit{etc.}\xspace}

\begin{document}

\title{CURLS: Causal Rule Learning for Subgroups with Significant Treatment Effect}

% \renewcommand{\thefootnote}{\fnsymbol{footnote}} % Adds the footnotes symbol

\author{Jiehui Zhou}
\email{zhoujiehui@zju.edu.cn}
\orcid{0000-0003-0709-775X}
\affiliation{%
  \institution{State Key Lab of CAD\&CG, Zhejiang University}
  \streetaddress{866 Yuhangtang Rd}
  \city{Hangzhou}
  \state{Zhejiang}
  \country{China}
  \postcode{310058}
}
\affiliation{%
  \institution{DAMO Academy, Alibaba Group}
  \streetaddress{Yuhang District-No. 969 Wenyi West Road}
  \city{Hangzhou}
  \state{Zhejiang}
  \country{China}
  \postcode{311121}
}

\author{Linxiao Yang}
\email{linxiao.ylx@alibaba-inc.com}
\orcid{0000-0001-9558-7163}
\affiliation{%
  \institution{DAMO Academy, Alibaba Group}
  \streetaddress{Yuhang District-No. 969 Wenyi West Road}
  \city{Hangzhou}
  \state{Zhejiang}
  \country{China}
  \postcode{311121}
}

\author{Xingyu Liu}
\email{liu_xingyu@zju.edu.cn}
\orcid{0009-0002-1009-4680}
\affiliation{%
  \institution{State Key Lab of CAD\&CG, Zhejiang University}
  \streetaddress{866 Yuhangtang Rd}
  \city{Hangzhou}
  \state{Zhejiang}
  \country{China}
  \postcode{310058}
}

\author{Xinyue Gu}
\email{guxinyue.gxy@alibaba-inc.com}
\orcid{0000-0002-6183-2295}
\affiliation{%
  \institution{DAMO Academy, Alibaba Group}
  \streetaddress{Yuhang District-No. 969 Wenyi West Road}
  \city{Hangzhou}
  \state{Zhejiang}
  \country{China}
  \postcode{311121}
}

\author{Liang Sun}
\authornotemark[1]
\email{liang.sun@alibaba-inc.com}
\orcid{0009-0002-5835-7259}
\affiliation{%
  \institution{DAMO Academy, Alibaba Group}
  \streetaddress{Yuhang District-No. 969 Wenyi West Road}
  \city{Hangzhou}
  \state{Zhejiang}
  \country{China}
  \postcode{311121}
}

\author{Wei Chen}
\email{chenvis@zju.edu.cn}
\orcid{0000-0002-8365-4741}
\affiliation{%
  \institution{State Key Lab of CAD\&CG, Zhejiang University}
  \streetaddress{866 Yuhangtang Rd}
  \city{Hangzhou}
  \state{Zhejiang}
  \country{China}
  \postcode{310058}
}

\authornote{Wei Chen and Liang Sun are corresponding authors.}

\renewcommand{\shortauthors}{Jiehui Zhou et al.}

\begin{abstract}
  In causal inference, estimating heterogeneous treatment effects (HTE) is critical for identifying how different subgroups respond to interventions, with broad applications in fields such as precision medicine and personalized advertising. Although HTE estimation methods aim to improve accuracy, how to provide explicit subgroup descriptions remains unclear, hindering data interpretation and strategic intervention management. In this paper, we propose \algname, a novel rule learning method leveraging HTE, which can effectively describe subgroups with significant treatment effects. Specifically, we frame causal rule learning as a discrete optimization problem, finely balancing treatment effect with variance and considering the rule interpretability. We design an iterative procedure based on the minorize-maximization algorithm and solve a submodular lower bound as an approximation for the original. Quantitative experiments and qualitative case studies verify that compared with state-of-the-art methods, \algname can find subgroups where the estimated and true effects are 16.1\% and 13.8\% higher and the variance is 12.0\% smaller, while maintaining similar or better estimation accuracy and rule interpretability. Code is available at \url{https://osf.io/zwp2k/}. 
  % \footnote{Code is available at \href{https://osf.io/zwp2k/?view_only=bb95c7a70eae40acb71cbadbbf9c8293}{https://osf.io/zwp2k/?view\_only=bb95c7a70eae40acb71cbadbbf9c8293}}
\end{abstract}







\maketitle

\input{sections/1_introduction}
\input{sections/2_related_work}
\input{sections/3_preliminary}
\input{sections/4_problem_formulation}
\input{sections/5_algorithm}
\input{sections/6_experiments}
\input{sections/7_discussion}
\input{sections/8_conclusion}

\begin{acks}
This work was supported by National Natural Science Foundation of China (62132017), Zhejiang Provincial Natural Science Foundation of China (LD24F020011) and Alibaba Group through Alibaba Research Intern Program.
\end{acks}

\newpage
\bibliographystyle{ACM-Reference-Format}
\bibliography{sample-base}

\appendix

\input{sections/9_appendix}

\end{document}

%% file: sections/1_introduction.tex
\section{Introduction}

Causal inference is a data analysis process aiming at conclusions about whether and to what extent treatments affect outcomes~\cite{peters2017elements}. Data heterogeneity needs to be taken into account when estimating treatment effects, as the effect of the same treatment often varies across subgroups. Discovering those subgroups with large effects and low variance (hereinafter referred to as significant treatment effect) compared to the overall population is widely used in domains such as healthcare~\cite{rothman2005causation}, marketing~\cite{varian2016causal}, and public administration~\cite{gangl2010causal}. For example, marketers would like to find customer groups where advertising is more effective in driving purchases. Since randomized controlled trials (RCTs), the gold standard for causal inference, are not always feasible due to cost or ethical concerns, there is a strong need to uncover subgroups with significant treatment effects from observational data.
% Thus, there is a strong demand for mining subgroups from observational data where treatments have significant effects on outcomes.

Existing causal models that consider the data heterogeneity, such as propensity score-based methods~\cite{Funk2011DoublyRE}, double machine learning~\cite{chernozhukov2018double}, meta-learners~\cite{kunzel2019metalearners}, entropy balancing~\cite{Hainmueller2012EntropyBF} and tree-based recursive partitioning~\cite{athey2016recursive, davis2017using}, can estimate the effect of the treatment on the outcome on subgroups or individuals. However, most of these methods try to reduce the confounding bias in the estimation rather than directly learning subgroups with significant effects. Researchers have also explored rule learning~\cite{Dash2018BooleanDR, NEURIPS2021_eaa32c96} and subgroup discovery~\cite{Grosskreutz2009OnSD, Leeuwen2012DiverseSS}, utilizing easy-to-understand rules to describe subgroups with intriguing patterns. Unfortunately, most rule learning methods are oriented towards correlations rather than causality, which may lead to imprecise estimations in selection bias-affected interventions. Thus, enhancing the treatment effect interpretability in the context of data heterogeneity is still underexplored.

\begin{figure}[htb]
  \centering
  \includegraphics[width=\linewidth]{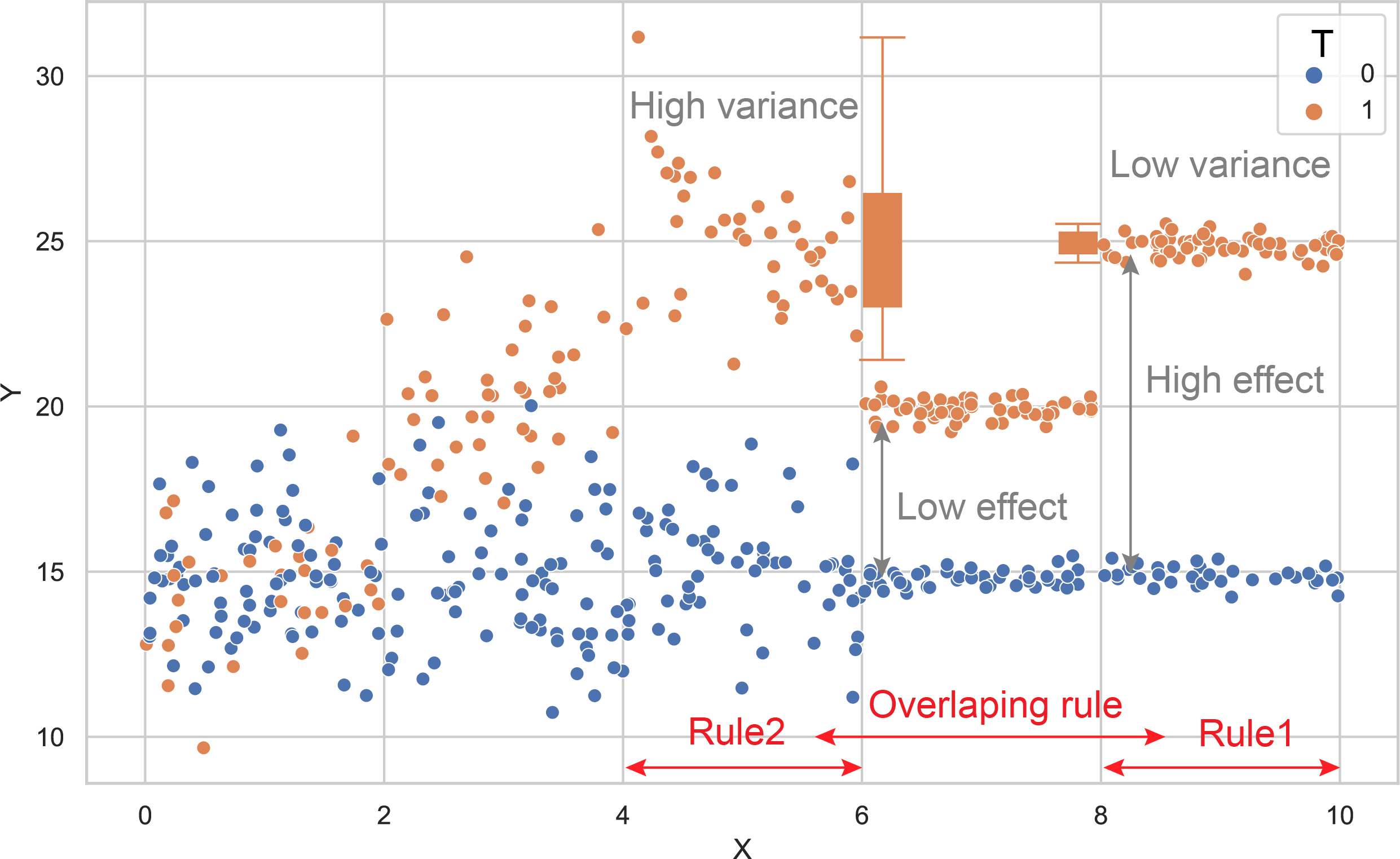}
  \caption{An illustrative toy example. There is only one covariate $X$, and the change in $Y$ can be informally thought of as the effect. The subgroup corresponding to Rule1 has a high effect and low variance, which the users expect to find.}
  \label{fig:toy}
\end{figure}

We combine the strong analytical power of heterogeneous treatment effect (HTE) estimation methods with the high interpretability of rule learning to facilitate the identification and interpretation of subgroups with significant treatment effects. However, two challenges must be addressed. First, the trade-off between multiple objectives and constraints is not trivial. Since treatment effect estimation is a statistical inference problem, in addition to requiring a large effect value, it is also necessary to ensure that the uncertainty of the estimates is small (as shown in Rule 1 in Fig.~\ref{fig:toy}). Also, the length and overlap of the rules are important constraints in order to facilitate user interpretation. Second, recognizing useful subgroups from a large number of potential candidates is difficult. Subgroups can be obtained by combining different attributes and values, which is often exponential and requires an efficient solution.

We propose \algname, a \underline{c}a\underline{u}sal \underline{r}ule \underline{l}earning method for identifying \underline{s}ubgroups with significant treatment effects. To address the first challenge, we formally define causal rules, which consist of subgroups described by conjunctive normal forms (CNFs) and the corresponding effects estimated by inverse probability weighting (IPW)~\cite{Hirano2000EfficientEO}. Then, mining subgroups with large treatment effects and low variance can be modeled as a discrete optimization problem. For the second challenge, we prove the existence of an approximate submodular lower bound for the optimization objective and design a solution based on the minorize-maximization (MM) algorithm and submodular optimization. Comprehensive quantitative experiments and qualitative case studies demonstrate the effectiveness of \algname. In summary, our contributions are as follows:

\begin{itemize}[noitemsep,topsep=0pt]
    \item We pioneer the incorporation of rule learning into causal inference, aiming to delineate subgroups with significant treatment effects through rule-based descriptions. Specifically, we formulate this as an optimization problem, considering the trade-off between effects and variance and rule set size and overlap constraints.
    
    \item We propose an efficient optimization algorithm that iteratively maximizes the submodular lower bound of the original problem, which cuts down the original exponential search space.

    \item We conduct both quantitative and qualitative experiments, demonstrating that \algname delivers not only extra rule-based interpretative capabilities for subgroups, but also enhances the precision of effect estimation with a smaller variance. Our method outperforms state-of-the-art algorithms in estimated and true effect strength (CATE) by 16.1\% and 13.8\%, respectively, and reduces variance by 12.0\%.
\end{itemize}

%% file: sections/2_related_work.tex
\section{Related Work}\label{sec:related:work}
In this section, we first review the algorithms related to HTE, then discuss the progress of rule learning, and finally summarize the work of subgroup discovery.

\subsection{Heterogeneous Treatment Effect Estimation}

Causal inference identifies the effect of treatment on outcome. However, treatment effects are often not ``one-size-fits-all''---they may vary across the population. Current HTE research is divided into conditional average treatment effect (CATE) and individual average treatment effect (ITE) by population level. Reviews~\cite{10.1145/3444944, 10.1145/3397269} provide detailed analyses on treatment effect estimation.

% ATE can be obtained by directly comparing the treatment and control groups in RCTs, where uplift modeling~\cite{DBLP:conf/icdm/RzepakowskiJ10} is the commonly used estimation method. However, in observational data, the distribution between the treatment and control groups may be inconsistent due to confounding variables, leading to biased results. Therefore, researchers have proposed various solutions. For example, Rosenbaum \etal~\cite{rosenbaum1983central} proposed a matching method to match each study subject with a subject having the same or similar covariate values, making the distributions of the treatment and control groups similar. Hirano \etal~\cite{hirano2003efficient} introduced inverse propensity weights in the estimation process to balance data with different degrees of representation. However, treatment effects may vary due to individual diversity, and simply adopting ATE may lead to poor outcomes~\cite{ling2022emulate}.

% HTE methods have been proposed for estimating the treatment effects of similar groups or individuals, \ie CATE and ITE.
CATE examines treatment effects on specific subgroups of the population, conditional on similar covariates, such as certain demographic characteristics. Tree-based methods~\cite{athey2016recursive, wager2018estimation, athey2019estimating} are widely used by dividing the covariate space into subspaces to maximize the treatment effects heterogeneity. For example, Causal Tree~\cite{athey2016recursive} uses part of the data to construct the tree and another part to estimate the treatment effect in each subspace, avoiding overfitting by cross-validation. To make the estimation more robust and smooth, Wager \etal~\cite{wager2018estimation} proposed Causal Forest, which aggregates the results of causal tree ensembles. The advantage of the tree model is its interpretability, which naturally provides subgroups of heterogeneous CATEs defined by root-to-leaf node paths.

ITE measures the difference in outcomes for individuals with or without receiving the treatment. Since only one outcome can be observed in the actual scenario, another potential outcome needs to be estimated. Depending on whether the treatment and control groups are estimated separately, existing methods can be categorized as single-model-based and multi-model-based. The former fits treatment effects with regression models. For example, Hill \etal~\cite{hill2011bayesian} use Bayesian additive regression trees to fit the outcome surface. The latter fits the treated and control groups separately and can achieve better performance when the difference between the outcomes of the two groups is significant. The base model can use off-the-shelf estimators, such as linear regression~\cite{cai2011analysis} or neural networks~\cite{johansson2016learning}. Although these models can be accurate in estimating effects with carefully tuned parameters, they are generally uninterpretable.

Previous work has focused on how to estimate effects more accurately, \ie to exclude confounding bias in the observational data. Instead, we aim to mine subgroups that have stronger effects with small variances. To this end, we utilize a propensity-score-based effect estimation method in our implementation and incorporate the ability of rules to characterize subgroups.

\subsection{Rule Learning}

Rules are simple logical structures of the form "IF P THEN Q". Since general rules are similar to the way humans think, rule learning is employed in prediction or classification scenarios that require high interpretability. The existing work can be broadly classified into pre-mining and uniform optimization approaches.

Most studies~\cite{Friedman2008PREDICTIVELV, Lakkaraju2016InterpretableDS, Wang2017ABF, pmlr-v108-mita20a, Zhang2020DiverseRS} adopted the two-stage paradigm consisting of rule generation (or rule pre-mining) and rule selection. First, rules are pre-mined through efficient algorithms such as decision trees and association rule mining to reduce the search space of rules significantly. In the second stage, appropriate rules are selected from the candidate rules to form an unordered rule set or an ordered rule list based on specific metrics (\eg classification accuracy). However, this separation paradigm can lead to sub-optimal results as important rules may be missed in the rule pre-mining stage, resulting in a loss of accuracy.

Recently, researchers have linked rule generation and selection in a single optimization framework for rule learning. For example, Dash \etal~\cite{Dash2018BooleanDR} formalized the rule set learning problem as an integer programming problem to balance classification accuracy and rule simplicity. Column generation (CG) algorithms were used to efficiently search for an exponential number of candidate clauses (conjunctive or disjunctive terms) without the need for heuristic rule mining. Yang \etal~\cite{NEURIPS2021_eaa32c96} approached rule set learning from the perspective of submodular optimization. They formulate the main problem as a task of selecting a subset from all possible rules, while the subproblem of searching for rules is mapped as another feature subset selection task. Since the objective function can be formulated as the difference between two submodular functions, it can be approximately solved by the optimization algorithm.

Most rule learning methods are used for classification tasks that solely examine correlations. However, correlation does not imply causation. A few researchers~\cite{Lee2020CausalRE, Li2015FromOS, wang2022causal, wu2023causal} have attempted to mine causal rules from data. For example, CRE~\cite{Lee2020CausalRE} and CRS~\cite{wang2022causal} are both two-stage methods, which first generates a pool of rules using random forest, FP-Growth, \etc, and then select a subset among them based on some criteria, such as stability selection regularization. Li \etal~\cite{Li2015FromOS} first mined the association rules from the data and then used a cohort study to test whether the association rules are causal or not. However, these methods lack a global optimization objective; therefore their results depend on the quality of the candidate rules developed in the first stage.

Unlike earlier methods, we mine causal rules from observational data from a unified optimization perspective. These rules represent those subgroups with large treatment effects and low variance.

\subsection{Subgroup Discovery}

Subgroup discovery (SD) is a descriptive data mining technique that identifies data subgroups with interesting patterns on specific targets. It differs slightly from rule learning that focuses on prediction/classification performance on upcoming data. A comprehensive study of SD is available in reviews~\cite{Herrera2011AnOO, AtzmuellerSubgroupD}.

Data subgroups can be represented using description languages such as attribute-value pairs and different logical forms (\eg conjunctions, disjunctions, inequalities, fuzzy logic, \etc). Subgroup interestingness can be measured using binary, nominal, or numerical targets. Certain post-processing methods have been applied to select diverse and less redundant subgroups. Due to the enormous number of potential subgroups, different search strategies, such as exhaustive and heuristic search, have been applied.

The exhaustive method~\cite{Wrobel1997AnAF, Atzmller2006SDMapA, Grosskreutz2008TightOE, Grosskreutz2009OnSD} searches all feasible subgroups. The naive exhaustive search may be time-consuming because the viable subgroup is exponential. Examples of strategies for reducing hypothesis space include optimistic estimate pruning, generalization-aware pruning, minimum support. SD-Map~\cite{Atzmller2006SDMapA} is a typical exhaustive SD method that extends the popular Frequent Pattern (FP) Growth-based association rule mining method, utilizing depth-first search to generate candidates. Piatetsky-Shapiro, unusualness, and binomial tests are utilized to determine precise and significant subgroups. The SD-Map*~\cite{Atzmller2009FastSD} is extended for use with binary, categorical, and continuous target variables.

Further studies~\cite{Gamberger2002ExpertGuidedSD, Lavra2004SubgroupDW, Leeuwen2012DiverseSS, Jess2007EvolutionaryFR, zitzler2001spea2} employed efficient heuristic methods. For example, DSSD~\cite{Leeuwen2012DiverseSS} is an SD algorithm based on beam search. The search usually starts with an initial solution and is then expanded to a certain number of candidate solutions. The best ones are retained for the next iteration until a stopping condition is reached. SDIGA~\cite{Jess2007EvolutionaryFR} is an evolutionary fuzzy rule induction algorithm. It facilitates the discovery of more general rules by allowing variables to take multiple values. Subgroups can be evaluated in terms of confidence, support, and unusualness.

SD is useful in many fields. For example, in medicine, it helps to discover high-risk groups for a certain disease~\cite{Lavra2005SubgroupDT}. During operation and maintenance, it helps troubleshoot and attribute anomalies in total KPI metrics to specific subgroups~\cite{Bhagwan2014AdtributorRD, Gu2020EfficientII}. In marketing, it helps to identify target customers of different brands~\cite{Lavra2004DecisionST}.

However, conventional SD methods usually overlook the treatment effect. Thus, this paper seeks to uncover subgroups with significant treatment effects, which requires different optimization objectives and evaluation criteria from prior SD methods.

%% file: sections/3_preliminary.tex
\section{Preliminaries}

In this section, we present some preliminaries about causal inference, submodular and supermodular functions.

\textit{Causal Inference}.
We introduce causal inference under the potential outcome (PO) framework~\cite{10.1093/oxfordhb/9780199286546.003.0011}. %These are basic terms.

\textbf{Unit}: A unit is an individual or object under study. A medical study unit may be a patient. The subscript ${}_{i}$ denotes the $i$-th unit.

\textbf{Treatment}: A treatment is an intervention or exposure that subjects to a unit. A new medicine or therapy could be used as a treatment in a medical study. Let $T$ indicate the treatment. $T=1$ units are the treatment group, while $T=0$ units are the control group. We assume one binary treatment for simplicity.

\textbf{Outcomes}: Outcomes are what would have happened under different treatments. Each unit has two potential outcomes: factual outcome and counterfactual outcome. Potential outcome with treatment value $t$ is $Y(T=t)$, also abbreviated as $Y(t)$.

% one for the treatment condition they actually received (factual outcome $Y^F$) and another for the condition they would have had if assigned to an alternative condition (counterfactual outcome $Y^{CF}$). The general health or survival of the patient may be potential outcomes in a medical study. The potential outcome with treatment value $t$ is denoted as $Y(T=t)$, also abbreviated as $Y(t)$.

% The outcome in the following sections refers to the observed outcome unless otherwise noted.

\textbf{Covariates}: Covariates are background variables that affect treatment assignment and outcome. Observational studies often control for covariates to mitigate confounding. Let $\mathbf{X}_i=(x_{i,1},\cdots,x_{i,d})$ represent covariates.

% Covariates are frequently employed to correct for confounding in observational studies. For example, patient demographic information such as age may influence medication use (treatment assignment) and blood pressure (outcome). Let $\mathbf{X}_i=(x_{i,1},\cdots,x_{i,d})$ represent covariates.

\textbf{Observational data}: Observational data refers to data collected without the researcher manipulating the environment or the subjects being studied. It differs from RCTs, which randomly assign treatment to each unit. The observational data containing $n$ units is denoted by $\mathcal{D} = \{(T_i,\mathbf{X}_i,Y_i)\}_{i=1}^n$.

\textbf{Treatment effect} refers to the impact of a treatment on an outcome. For observational data, Inverse Probability Weighting (IPW)~\cite{Hirano2000EfficientEO} is proposed, which assigns appropriate weights $ w_i=\frac{T_i}{e_i} + \frac{1-T_i}{1-e_i}$ to each unit based on propensity score $e_i$ to balance covariates distribution in the treatment and control groups, thereby simulating RCTs. Then, the normalized weighted average of the factual outcomes for the treatment and control groups can be calculated to estimate treatment effects~\cite{10.1162/003465304323023651}:

\begin{equation}
    \tau = \dfrac{\sum_{i;T_i=1} w_iY_i}{\sum_{i;T_i=1} w_i} - \dfrac{\sum_{i;T_i=0}w_iY_i}{\sum_{i;T_i=0}w_i}. 
\end{equation}

When the data satisfy the three causal assumptions (unconfoundedness, positivity and stable unit treatment value), it is shown that the adjustment of the scalar propensity score removes the bias due to the observed covariates~\cite{Rosenbaum1983TheCR}.

\textit{Submodular and supermodular functions}.
A submodular function is a set function with special properties. Its domain is a family of subsets of a given set. The output value is some measure of the subset. The inputs and outputs satisfy the relationship of diminishing returns, \ie the additional benefit of adding the set to the inputs declines. The supermodular function is the opposite of the submodular function, which satisfies the increasing returns. Formally, for a set function $f: 2^\Omega \rightarrow \mathbb{R}$, it is submodular if:

\begin{equation*}
    \forall A \subseteq B \subseteq \Omega \, \text{and} \, v \in \Omega \setminus B, f(A \cup \{v\}) - f(A) \geq f(B\cup\{v\}) - f(B). 
\end{equation*}

Like the convexity in continuous optimization, submodularity is a good property in discrete optimization, making it suitable for many applications, such as approximation algorithms, game theory, automatic summarization, and feature selection~\cite{Lin2011ACO, krause2008beyond}.

%% file: sections/4_problem_formulation.tex
\section{Problem Formulation}

This section introduces the formalization of the causal rule learning problem. The final optimization problem is presented in Eq.~\eqref{eq:opt}.

Without loss of generality, we consider observational data whose covariates are binary and outcome is a positive number, \ie $\mathbf{X}_i=(x_{i,1},\cdots,x_{i,d})\in\{0,1\}^d, Y \in \mathbb{R}_{>0}$. Categorical variables can be binarized by one-hot encoding, and numerical variables can be converted to binary by bucketing strategy. We determine the optimal number of bins (4-20) using 5-fold cross-validation and select the best-performing parameter for the final model. Negative outcomes can be made positive by adding an offset.

Formally, given the observational data $\mathcal{D}$, we aim to learn interpretable causal rules from it.
% As shown in the figure xx, 
A \textbf{causal rule} $\mathcal{R}: \bm{\alpha} \Rightarrow \tau$ contains the antecedent $\bm{\alpha}$ and the consequent $\tau$.

A \textbf{antecedent} $\bm{\alpha}$ is the condition of the rule, expressed as the conjunctive normal form (CNF) of a series of atoms $\bigwedge_{j \in \Gamma} x_{j}$, \eg "age > 25 AND job != teacher". $\Gamma$ is the covariate indices used in the antecedent, which is a subset of the indices of all binary covariates, \ie $\Gamma \in 2^{[d]}$, where $[d] = \{1, \cdots, d\}$ and $2^{[d]}$ means the power set of $[d]$. The atom $x_j$ is the smallest interpretable element. We create the negation of covariate $\neg x_j$ to increase the expressiveness. The mapping from a $\mathcal{R}$ to a CNF is given by $\bm{\alpha}_{\mathcal{R}}(\mathbf{X}_i) = \bigwedge_{j \in \Gamma_{\mathcal{R}}} x_{i,j}$. For brevity, we also refer to it as $\bigwedge_{j \in \mathcal{R}} x_{i,j}$. When $\bm{\alpha}_{\mathcal{R}}(\mathbf{X}_i)$ is true, the $i$-th unit is \textbf{covered} by the rule $\mathcal{R}$.

The \textbf{consequent} $\tau$ is the prediction result of the rule, indicating the estimated treatment effect for the data covered by the rule. Define $\mathcal{D}_{\mathcal{R}}$ to denote the covered data, $\mathcal{D}^+$ to denote the data that received the treatment ($T=1$), and $\mathcal{D}^-$ to denote the data that did not receive the treatment ($T=0$). Define $\mathcal{D}_{\mathcal{R}}^+ = \{ i | i \in \mathcal{D}^+ \wedge \bm{\alpha}_{\mathcal{R}}(\mathbf{X}_i) = 1 \}$ denotes the units in the treatment group that are covered by the rule $\mathcal{R}$, $\mathcal{D}_{\mathcal{R}}^- = \{ i | i \in \mathcal{D}^- \wedge \bm{\alpha}_{\mathcal{R}}(\mathbf{X} _i) = 1 \}$ denotes the units in the control group that are covered by the rule $\mathcal{R}$. Therefore, the treatment effect of rule $\mathcal{R}$ can be represented as:

\begin{equation}
    \tau_{\mathcal{R}} = \dfrac{\sum_{i \in \mathcal{D}_{\mathcal{R}}^+} w_iY_i}{\sum_{i \in \mathcal{D}_{\mathcal{R}}^+} w_i} - \dfrac{\sum_{i \in \mathcal{D}_{\mathcal{R}}^-}w_iY_i}{\sum_{i \in \mathcal{D}_{\mathcal{R}}^-}w_i} = \dfrac{Q_1}{Q_2} - \dfrac{Q_3}{Q_4},
\end{equation}
where $Q_1 = \sum_{i \in \mathcal{D}_{\mathcal{R}}^+} w_iY_i$, $Q_2 = \sum_{i \in \mathcal{D}_{\mathcal{R}}^+} w_i$, $Q_3=\sum_{i \in \mathcal{D}_{\mathcal{R}}^-}w_iY_i$, $Q_4=\sum_{i \in \mathcal{D}_{\mathcal{R}}^-}w_i$.

The \textbf{causal rule set} $\mathcal{S} = \{\mathcal{R}_1, \cdots, \mathcal{R}_k\}$ contains multiple rules. A rule set covers a unit if $\bm{\alpha}_{\mathcal{S}}(\mathbf{X}_i) = \bigvee_{\mathcal{R} \in \mathcal{S}} \bigwedge_{j \in \mathcal{R}} x_{i, j}$ is true. If a unit is covered by more than one rule, we take the average of the effects as the estimated treatment effect for that unit. However, an interpretable ruleset should minimize rule overlap.

Since obtaining treatment effects is a statistical estimation problem, it is important to consider the uncertainty of the treatment effect, which can be measured by the outcome variance of the treatment group, defined as:

\begin{equation}
    \sigma_{\mathcal{R}}^2=\dfrac{\sum_{i \in \mathcal{D}_{\mathcal{R}}^+} w_i(Y_i - \overline Y_w)^2}{\sum_{i \in \mathcal{D}_{\mathcal{R}}^+} w_i},
\end{equation}

For the effect, we want $\frac{Q_1}{Q_2}$ to be large, $\frac{Q_3}{Q_4}$ to be small, and the variance $\sigma$ is also small. To facilitate the optimization, we take a log for the ratio. Then we get the \textbf{objective function} $f(\mathcal{R}) = \log Q_1 + \log Q_4 - \log Q_2 - \log Q_3 - \lambda \log \sigma_{\mathcal{R}}^2$, which is the profit of a rule, and $\lambda$ is a coefficient that adjusts for the trade-off between treatment effect and variance. Therefore, to learn the causal rule set from observational data, we consider solving the following optimization problem: 

\begin{equation}
\begin{aligned}
\label{eq:opt}
\max_{\mathcal{S}} & \sum_{\mathcal{R} \in \mathcal{S}}
f(\mathcal{R}) \\
\text {s.t.} & |\mathcal{S}| \leq K \\
& |\mathcal{R}| \leq L.
\end{aligned}
\end{equation}
To ensure interpretability, $|\mathcal{S}| \leq K$ restricts the number of rules in the rule set to be no more than $K$, and $|\mathcal{R}| \leq L$ limits the antecedent length of the rules to be no more than $L$.

%% file: sections/5_algorithm.tex
\section{Algorithm}

In this section, we introduce the proposed algorithm \algname for solving the optimization problem in Eq.~\eqref{eq:opt}. 

\begin{figure*}[htp!]
  \centering
  \includegraphics[width=2.1\columnwidth]{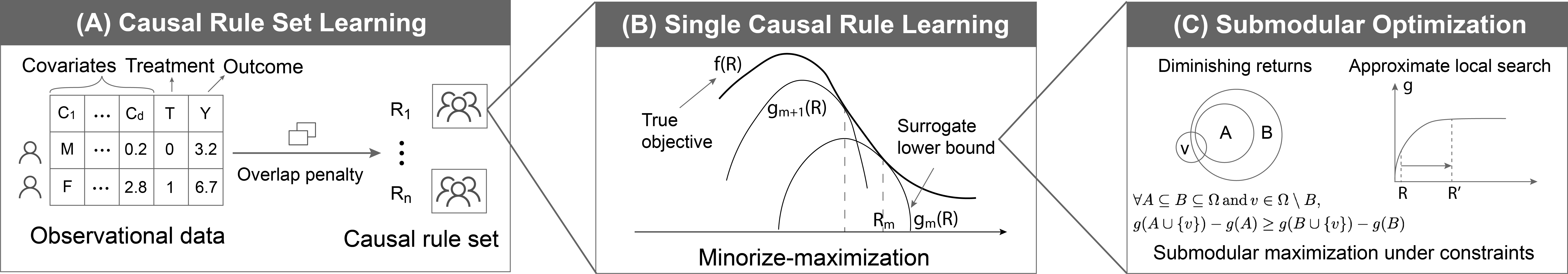}
  \caption{Illustration of the proposed algorithm. (A) A causal rule set is learned from the observational data, and overlap penalities are applied to minimize the case where a unit is covered by multiple rules. (B) A single causal rule is solved by the MM framework, and the rule is improved by iteratively optimizing the surrogate lower bound of the original objective. (C) We prove an surrogate lower bound with submodular properties, allowing us to efficiently solve the surrogate optimization problem using efficient submodular optimization.}
  \label{fig:workflow}
\end{figure*}

As shown in Fig.~\ref{fig:workflow}, to construct a precise causal rule set, we employ an iterative framework (Sec. \ref{alg:rule-set}), balancing constraints such as set size and antecedent length. This methodical approach allows us to sequentially select the most fitting rule, ensuring a coherent and optimized causal rule set. Central to our strategy is the optimization of a set function, which maps covariates to treatment effects, requiring a nuanced balance between computational feasibility and accuracy. We address this challenge by crafting an approximate submodular lower bound for the objective function, a strategic choice that simplifies the optimization process while maintaining solution quality. Leveraging the minorize-maximization (MM) procedure (Sec. \ref{alg:mm}) and submodular optimization(Sec. \ref{alg:sub}), we efficiently derive each rule's antecedents and consequents, resulting in significant and interpretable causal rules.

% To obtain the causal rule set, we adopt an iterative solving framework. Taking into account constraints such as overlap, rule set size and rule antecedent length, we find the best rule one at a time and then add it to the rule set (Sec. \ref{alg:rule-set}). For solving a rule, it is a problem of optimizing a set function (from a subset of covariates to treatment effect value). We construct the approximate submodular lower bound of the objective function, and use the minorize-maximization (MM) procedure (Sec. \ref{alg:mm}) and the submodular optimization method (Sec. \ref{alg:sub}) to obtain the antecedent and consequent corresponding to the rule.

\subsection{Causal Rule Set Learning} \label{alg:rule-set}

Directly optimizing the rule set is not a trivial problem. Typical correlation rule set learning algorithms usually adopt sequential covering paradigms~\cite{Cohen1995FastER}, that is, removing data covered by previous rules and learning a new rule. However, this can easily result in overlapping rules, thus affecting the interpretability of the rule set. To solve this problem, instead of removing the covered data, we explicitly introduce a penalty for overlapping data in the iterative process, thus increasing the diversity of the rule set. 

\textbf{Overlap Penalty.} Since the purpose of the causal rule is to cover those units that have a strongly positive outcome after receiving treatment, we can set the weighted outcome of the covered units belonging to the treatment group to a smaller value $\epsilon$, \ie

\begin{equation}
    Q_1(\mathcal{R}) = \sum\nolimits_{i \in \mathcal{D}_{\mathcal{R}}^+ \setminus \mathcal{D}_{\mathcal{S}}} w_iY_i + \sum\nolimits_{i \in \mathcal{D}_{\mathcal{R}}^+ \cap \mathcal{D}_{\mathcal{S}}} \epsilon. 
\end{equation}
Therefore, if the new rule $\mathcal{R}$ searches for units that have been covered by the current rule set $\mathcal{S}$, its estimated effect will be low, and thus its probability of being selected during the optimization process will decrease. The overall causal rule set learning process is shown in Alg. \ref{alg1}.

\begin{algorithm}
    \caption{Causal rule set learning}
    \label{alg1}
\begin{algorithmic}[1]
    \STATE {\bfseries Input:} Training data $\mathcal{D} = \{(T_i,\mathbf{X}_i,Y_i)\}_{i=1}^n$,
    hyperparameters $\lambda$, $K$, and $L$

    \STATE Initialize $\mathcal{S} \gets \emptyset$
    \FOR{$k = 1$ {\bfseries to} $K$}
        \STATE Solve $\mathcal{R}^{\star} \gets \arg\max_{\mathcal{R}} f(\mathcal{R})$ \COMMENT{See Sec. \ref{alg:mm}}
        \IF{$f(\mathcal{R}^{\star}) > 0$}
            \STATE $\mathcal{S} \gets \mathcal{S} \cup \{ \mathcal{R}^{\star} \} $
            \STATE Change weighted outcome to $\epsilon$ for covered units
        \ENDIF
    \ENDFOR
    \STATE {\bfseries Output:} $\mathcal{S}$
\end{algorithmic}
\end{algorithm}

\subsection{MM Procedure} \label{alg:mm}

For a single rule, its maximization objective function is $f(\mathcal{R})$. However, the complexity of $f(\mathcal{R})$ makes it difficult for traditional optimization algorithms to handle it directly. To this end, we propose to use the MM procedure. It is an iterative optimization method that, instead of finding the optimal solution to the original objective function $f(\mathcal{R})$, first finds an easy-to-optimize surrogate function $g(\mathcal{R})$ that approximates the original one (see Sec. \ref{alg:sub} for detail). The solution of the surrogate function makes the optimal solution of $g(\mathcal{R})$ close to the optimal solution of $f(\mathcal{R})$. In each iteration, a new surrogate function for the next iteration is constructed based on the current solution. Mathematically, the solution can converge to the optimal solution to the original optimization problem~\cite{MM:survey}.
%%%Through many iterations, a solution that is closer and closer to the optimal solution of the original objective function can be obtained.

Formally, taking the minorize-maximization version, $f(\mathcal{R})$ is the original objective function to be maximized.
At the $m-$th ($m=0,1,\dots$) step of MM, the objective function $f(\mathcal{R})$ can be replaced by a surrogate function $g_m(\mathcal{R})$ if the following conditions are satisfied:

\begin{equation}
    \begin{aligned}
    & g_m(\mathcal{R}) \le f(\mathcal{R}) \quad \forall \mathcal{R}  \\
    & g_m(\mathcal{R}_m) = f(\mathcal{R}_m).
    \end{aligned}
\end{equation}
Formmally, we summarize the steps of the MM procedure in Alg.~\ref{alg2}.

\begin{algorithm}
    \caption{Single causal rule optimization}
    \label{alg2}
\begin{algorithmic}[1]
    % \STATE {\bfseries Input:} Current solution $\mathcal{R}$
    \STATE $m = 0$
    \STATE Initialize $\mathcal{R}_m$
    \WHILE{\TRUE}
        % \STATE $\mathcal{R}^{\prime} \gets \mathcal{R}_m$
        \STATE Construct $g_m(\mathcal{R})$ \COMMENT{See Sec. \ref{alg:sub}}
        \STATE $\mathcal{R}_{m+1} = \arg \max_{\mathcal{R}} g_m(\mathcal{R})$
        \STATE \algorithmicif \ $\mathcal{R}_{m+1} = \mathcal{R}_m$ \algorithmicthen \ $\mathcal{R}^{\star}=\mathcal{R}_m$ {\bfseries break}
        \algorithmicend \ \algorithmicif
        \STATE $m = m + 1$
    \ENDWHILE
    \STATE {\bfseries Output:} $\mathcal{R}^{\star}$
\end{algorithmic}
\end{algorithm}

\subsection{Submodular Lower Bound Optimization} \label{alg:sub}

Here, we introduce how to construct a submodular approximation lower bound of the original objective function and the corresponding optimization method. Specifically, we aim to develop a rule $\mathcal{R}^{\star}$ that maximizing $f(\mathcal{R})$. First, we introduce the following inequality.
%%%e get the following result.

\begin{proposition}
$\sigma_{\mathcal{R}}^2 \le \frac{\sum_{i \in \mathcal{D}_{\mathcal{R}}^+} w_i(Y_i - \mu_{(m)})^2}{\sum_{i \in \mathcal{D}_{\mathcal{R}}^+} w_i}$, where $\mu_{(m)}$ is the weighted mean of outcome of the previous step in the MM procedure.
\end{proposition}

\begin{proof}
    $\sigma_{\mathcal{R}}^2$ is the weighted variance. According to the definition, it can also be written as: 
    $\sigma_{\mathcal{R}}^2=\mathbb{E}_w[Y_i^2] - (\mathbb{E}_w[Y_i])^2 = \frac{\sum_{i \in \mathcal{D}_{\mathcal{R}}^+} w_iY_i^2}{\sum_{i \in \mathcal{D}_{\mathcal{R}}^+} w_i} - (\frac{\sum_{i \in \mathcal{D}_{\mathcal{R}}^+} w_iY_i}{\sum_{i \in \mathcal{D}_{\mathcal{R}}^+} w_i})^2$. We perform Taylor expansion of the latter term, so that $\sigma_R^2 \le \frac{\sum_{i \in \mathcal{D}_{\mathcal{R}}^+} w_iY_i^2}{\sum_{i \in \mathcal{D}_{\mathcal{R}}^+} w_i} - (2\mu_{(m)}\frac{\sum_{i \in \mathcal{D}_{\mathcal{R}}^+} w_iY_i}{\sum_{i \in \mathcal{D}_{\mathcal{R}}^+} w_i} - \mu_{(m)}^2) = \\ \frac{\sum_{i \in \mathcal{D}_{\mathcal{R}}^+}w_i(Y_i^2 - 2\mu_{(t)}Y_i + \mu_{(m)}^2)}{\sum_{i \in \mathcal{D}_{\mathcal{R}}^+} w_i} = \frac{\sum_{i \in \mathcal{D}_{\mathcal{R}}^+} w_i(Y_i - \mu_{(m)})^2}{\sum_{i \in \mathcal{D}_{\mathcal{R}}^+} w_i}$.
\end{proof}

Define $Q_5 = \sum_{i \in \mathcal{D}_{\mathcal{R}}^+} w_i(Y_i - \mu_{(m)})^2$, $Q_6 = \sum_{i \in \mathcal{D}_{\mathcal{R}}^+} w_i = Q_3$, then $f(\mathcal{R}) \ge \log Q_1 + \log Q_4 - \log Q_3 - \log Q_4 - \lambda (\log \frac{Q_5}{Q_6}) = \log Q_1 + \log Q_4 + \lambda \log Q_6 - \log Q_2 - \log Q_3 - \lambda \log Q_5$.

Each atom (covariate) of the antecedent in the rule corresponds to a part of the unit, and the unit corresponding to the entire rule is the intersection of the units corresponding to these atoms. Based on the formula for $Q$, $Q$ can be viewed as the set function. Taking $Q_1 = \sum_{i \in \mathcal{D}_{\mathcal{R}}^+} w_iY_i$ as an example, its corresponding units is $\mathcal{D}_{\mathcal{R}}^+$, and the corresponding value is the sum of the weighted outcome of these units. Then, we have the following property for $Q$ functions. 

\begin{proposition}
$Q$ functions are supermodular.
\end{proposition}

\begin{proof}
    Taking $Q_1$ as an example, $\mathcal{D}_{\mathcal{R}}^+ = \{ i | i \in \mathcal{D}^+ \wedge \bm{\alpha}_{\mathcal{R}}(\mathbf{X}_i) = 1 \} = \{ i | i \in \mathcal{D}^+ \wedge (\wedge_{j \in \mathcal{R}} x_{i,j}) = 1 \}$. Thus $Q_1$ can be regarded as the weighted outcome sum of set $|\mathcal{D}^+ \cap (\cap_{j \in \mathcal{R}} x_j) |$. We can rewrite it as $|\mathcal{D}^+| - |\mathcal{D}^+ \cap (\overline{\cap_{j \in \mathcal{R}} x_j})| = |\mathcal{D}^+| - |\mathcal{D}^+ \cap (\cup_{j \in \mathcal{R}} \overline{x_j})| = |\mathcal{D}^+| - |\cup_{j \in \mathcal{R}}(\mathcal{D}^+ \cap \overline{x_j})|$. The latter term is the union of sets (coverage functions), which is a well-known submodular function, so $Q_1$ is a supermodular function. Similarly, it can be shown that other $Q$ functions are also supermodular.
\end{proof}

For the supermodular function $Q: 2^V \rightarrow \mathbb{R}_{\ge 0}$, where $V=[d]$ is the universal set. The following modular functions gives two tight lower bounds approximating $Q$ at $\mathcal{R}_m$~\cite{Nemhauser1978AnAO}:

% \begin{equation}
% \begin{split}
%     b^{1}_{Q,\mathcal{R}_m}(\mathcal{R}) & = Q(\mathcal{R}_m) - \sum_{j \in \mathcal{R}_m \setminus \mathcal{R}} Q(j | \mathcal{R}_m \setminus \{j\}) \\ & + \sum_{j \in \mathcal{R} \setminus \mathcal{R}_m} Q(j | \emptyset) \leq Q(\mathcal{R}), \forall \mathcal{R} \\
%     b^{2}_{Q,\mathcal{R}_m}(\mathcal{R}) & = Q(\mathcal{R}_m) - \sum_{j \in \mathcal{R}_m \setminus \mathcal{R}} Q(j | V \setminus \{j\}) \\
%     & + \sum_{j \in \mathcal{R} \setminus \mathcal{R}_m} Q(j | \mathcal{R}_m) \leq Q(\mathcal{R}), \forall \mathcal{R},
% \end{split}
% \end{equation}

\begin{equation}
\begin{split}
    b^{1}_{Q,\mathcal{R}_m}(\mathcal{R}) & = Q(\mathcal{R}_m) - \sum\nolimits_{j \in \mathcal{R}_m \setminus \mathcal{R}} Q(j | \mathcal{R}_m \setminus \{j\}) \\ & + \sum\nolimits_{j \in \mathcal{R} \setminus \mathcal{R}_m} Q(j | \emptyset) \leq Q(\mathcal{R}), \forall \mathcal{R} \subseteq V \\
    b^{2}_{Q,\mathcal{R}_m}(\mathcal{R}) & = Q(\mathcal{R}_m) - \sum\nolimits_{j \in \mathcal{R}_m \setminus \mathcal{R}} Q(j | V \setminus \{j\}) \\
    & + \sum\nolimits_{j \in \mathcal{R} \setminus \mathcal{R}_m} Q(j | \mathcal{R}_m) \leq Q(\mathcal{R}), \forall \mathcal{R} \subseteq V,
\end{split}
\end{equation}
where $Q(A|B) = Q(A \cup B) - Q(B)$ denotes the marginal gain from adding $A$ to $B$. For ease of expression, we define

\begin{equation}
    b_{Q,\mathcal{R}_m}(\mathcal{R}) = \max (b^{1}_{Q,\mathcal{R}_m}(\mathcal{R}), b^{2}_{Q,\mathcal{R}_m}(\mathcal{R})). 
\end{equation}

Then we have the following result: 

\begin{proposition} The function $g(\mathcal{R})$ defined below is a submodular lower bound of $f(\mathcal{R})$:
\label{proposition3}
\begin{align*}
g(\mathcal{R}) =& \log b_{Q_1,\mathcal{R}_m}(\mathcal{R}) + \log b_{Q_4,\mathcal{R}_m}(\mathcal{R}) + \lambda \log b_{Q_6,\mathcal{R}_m}(\mathcal{R}) \\
&- (\log Q_{2, \mathcal{R}_m} + \frac{Q_{2, \mathcal{R}} - Q_{2,\mathcal{R}_m}}{Q_{2,\mathcal{R}_m}}) - (\log Q_{3, \mathcal{R}_m} + \frac{Q_{3, \mathcal{R}} - Q_{3,\mathcal{R}_m}}{Q_{3,\mathcal{R}_m}}) \\
&- \lambda (\log Q_{5, \mathcal{R}_m} + \frac{Q_{5, \mathcal{R}} - Q_{5,\mathcal{R}_m}}{Q_{5,\mathcal{R}_m}}).
\end{align*}
\end{proposition}

\begin{proof}
    Since $b_{Q,\mathcal{R}_m}(\mathcal{R})$ is a modular function, then $\log b_{Q,\mathcal{R}_m}(\mathcal{R})$ is a submodular function. The first-order Taylor expansion of $\log Q$ is $\log Q_{\mathcal{R}_m} + \frac{Q_{\mathcal{R}} - Q_{\mathcal{R}_m}}{Q_{\mathcal{R}_m}}$, where $Q_{\mathcal{R}_m}$ are constants, so $-Q_{\mathcal{R}}$ is a submodular function. Thus, the $g(\mathcal{R})$ is a submodular function.
\end{proof}

For the submodular lower bound $g(\mathcal{R})$, there exists an approximate local search algorithm~\cite{Lee2009NonmonotoneSM} that approaches the optimum by continuously performing local improvements. Specifically, starting from the initial rule, $g(\mathcal{R})$ is gradually maximized by local operations, including adding, removing, or replacing covariates. Formally, the detailed algorithm procedures are summarized in Alg.~\ref{alg3}.

\begin{algorithm}
    \caption{Submodular lower bound optimization}
    \label{alg3}
\begin{algorithmic}[1]
    \STATE {\bfseries Input:} Current rule $\mathcal{R}$
    \WHILE{\TRUE}
        \STATE $\mathcal{R}^{\prime} \gets \mathcal{R}$

        \STATE \algorithmicwhile{\ $\exists j \in [d] \setminus \mathcal{R}$ s.t. $g(j | \mathcal{R}) > 0$}
                \algorithmicdo \ $\mathcal{R} \gets \mathcal{R} \cup \{j\}$ \algorithmicend \ \algorithmicwhile
        
        \STATE \algorithmicwhile{\ $\exists j \in \mathcal{R}$ s.t. $g(j | \mathcal{R} \setminus \{ j \}) \leq 0$}
            \algorithmicdo \ $\mathcal{R} \gets \mathcal{R} \setminus \{j\}$ \algorithmicend \ \algorithmicwhile
        
        \STATE \algorithmicwhile{\ $\exists i \in \mathcal{R}, j \in [d] \setminus \mathcal{R}$ s.t. $g(j | \mathcal{R} \setminus \{ i \}) > 0$}
            \algorithmicdo \ $\mathcal{R} \gets (\mathcal{R} \setminus \{i\}) \cup \{ j \}$ \algorithmicend \ \algorithmicwhile

        \STATE \algorithmicif \ $\mathcal{R} = \mathcal{R}^{\prime}$ \algorithmicthen \ {\bfseries break}
        \algorithmicend \ \algorithmicif
    \ENDWHILE
    \STATE {\bfseries Output:} $\mathcal{R}$
\end{algorithmic}
\end{algorithm}

%% file: sections/6_experiments.tex
\section{Evaluation}
We present detailed experimental evaluation of \algname, including quantitative experiments and qualitative case studies.

\subsection{Quantitative Experiments}

The quantitative experiments aim to evaluate the efficacy of \algname in identifying significant treatment effects in subgroups. Since real-world datasets lack the groundtruth of CATE, which affects the calculation of evaluation metrics, we compare \algname with various baselines on synthetic and semi-synthetic datasets.

\textbf{Datasets}. For synthetic data, following the settings in~\cite{athey2016recursive, Wu2023StableEO}, we sampled units under the assumption of unconfoundedness, where the covariates are generated from the following distribution:
\begin{equation}
    \begin{aligned}
    & X_{1}, \cdots, X_{i} \sim \text{Categorical}(\{A, B, C, D, E\}), \\
    & X_{i+1}, \cdots, X_{d} \sim \text{Normal}(0, 1).
    \end{aligned}
\end{equation}

The treatment $T$ is generated according to a Bernoulli distribution, where the probability of $T=1$ is given by the sigmod function with respect to $X$. This simulates the non-randomness of treatment assignment in the observational data. Categorical variables are converted to one-hot encoding for calculation. Formally, we have
%\begin{equation}
%    \begin{aligned}
%    & f(X)=\sigma(\langle X, \beta\rangle + \eta), \;\eta \sim \text{Uniform}(-1, 1), \beta \sim \text{Uniform}(0, b)^{|X|}, \\
%    & T \sim \text{Bernoulli}(f(X)). 
%    \end{aligned}
%\end{equation}

\begin{equation}
    \begin{aligned}
    f(X)&=\sigma(\langle X, \beta\rangle + \eta), \\
    \eta &\sim \text{Uniform}(-1, 1), \\
    \beta &\sim \text{Uniform}(0, b)^{|X|}, \\
    T &\sim \text{Bernoulli}(f(X)). 
    \end{aligned}
\end{equation}

The treatment effect $\text{TE}$ and the outcome $Y$ is generated by the following formula. An offset is added to $Y$ to ensure that $Y$ is positive. That is, 

\begin{equation}
    \begin{aligned}
    TE &= \langle X, \alpha \rangle, \alpha \sim \text{Uniform}(0,2)^{|X|},   \\
    Y &=  T \cdot \text{TE} + \langle X, \gamma\rangle + Y_{\text{offset}} + \epsilon, \\
    Y_{\text{offset}} &= \max(0, -Y_{\min}), \\
    \epsilon &\sim \text{Uniform}(-1, 1), 
    \gamma \sim \text{Uniform}(0,1)^{|X|}.
    \end{aligned}
\end{equation}

We also collected the famous semi-synthetic dataset IHDP~\footnote{\url{https://github.com/AMLab-Amsterdam/CEVAE/tree/master/datasets/IHDP}}, which is constructed from the infant health and development program. The detail information of the datasets is shown in Table~\ref{tab:data_sta}.

\begin{table}[htb]
\centering
\caption{
Dataset statistics.
}
\vspace{-3mm}
{
\begin{tabular}{lcccc}
\hline
Dataset & \#Units & \#Categorical\_cov & \#Numerical\_cov & b  \\ \hline
Syn-data1 & 3000 & 5 & 5 & 0.6 \\
Syn-data2 & 3000 & 5 & 10 & 0.5 \\
Syn-data3 & 4000 & 5 & 15 & 0.3 \\ 
IHDP & 7470 & 19 & 6 & / \\
\hline
\end{tabular}
}
\label{tab:data_sta}
\vspace{-4mm}
\end{table}

\textbf{Baselines}. We compare the proposed algorithm \algname with two groups of algorithms. The first group is the popular heterogeneous treatment effect estimation algorithms: (1) Causal Tree (CT)~\cite{athey2016recursive}; (2) Causal Forest (CF)~\cite{wager2018estimation}; and (3) Causal Rule Ensemble (CRE)~\cite{Lee2020CausalRE}. The second group is the correlation rule learning and subgroup discovery algorithms: (1) BRCG~\cite{Dash2018BooleanDR}; (2) Decision Tree (DT)~\cite{Breiman2017PointsOS}; (3) Pysubgroup (PYS)~\cite{lemmerich2018pysubgroup}. In the first group, CRE can explicitly obtain the antecedent and treatment effect of the rule. For CT and CF, it can be considered that the path from the root to the leaf nodes in the tree structure is the antecedent of the causal rule, and the CATE value of the leaf node is the effect corresponding to the rule. The second group of methods can only get the correlation rules. In order to adapt to the setting of causal rule learning, we add a post-processing step. CATE is calculated on the data covered by each rule via the normalized IPW method~\cite{10.1162/003465304323023651}. For the consistency of comparison, the rules with the top $K$ effect values in the baselines are taken out and compared with \algname.

\begin{table*}[htb]
    \centering
    \caption{5-fold average performance metrics for different rules. Numbers in parentheses represent standard deviations. (The CRE results on IHDP are missing because it cannot find any rules, and some standard deviations of BRCG are nan since sometimes only one fold can obtain a rule that meet the requirements.)}
    \vspace{-2mm}
    \begin{adjustbox}{width=\textwidth}
        \begin{tabular}{
            l
            |l
            |S[table-format=2.2,table-space-text-post=\tiny{(3.30)}]
            S[table-format=2.2,table-space-text-post=\tiny{(3.30)}]
            |S[table-format=2.2,table-space-text-post=\tiny{(3.30)}]
            S[table-format=2.2,table-space-text-post=\tiny{(3.30)}]
            |S[table-format=2.2,table-space-text-post=\tiny{(3.30)}]
            S[table-format=2.2,table-space-text-post=\tiny{(3.30)}]
            |S[table-format=2.2,table-space-text-post=\tiny{(3.30)}]
            S[table-format=2.2,table-space-text-post=\tiny{(3.30)}]
            |S[table-format=2.2,table-space-text-post=\tiny{(3.30)}]
            S[table-format=2.2,table-space-text-post=\tiny{(3.30)}]
            |S[table-format=2.2,table-space-text-post=\tiny{(3.30)}]
            S[table-format=2.2,table-space-text-post=\tiny{(3.30)}]
            |S[table-format=2.2,table-space-text-post=\tiny{(3.30)}]
            S[table-format=2.2,table-space-text-post=\tiny{(3.30)}]
        }
        \toprule
        \multirow{2}{*}{Dataset} & \multirow{2}{*}{Metrics} & \multicolumn{2}{c|}{CURLS} & \multicolumn{2}{c|}{CT} & \multicolumn{2}{c|}{CF} & \multicolumn{2}{c|}{CRE} & \multicolumn{2}{c|}{DT} & \multicolumn{2}{c|}{PYS} & \multicolumn{2}{c}{BRCG} \\
                                 &                          & {Rule1} & {Rule2}         & {Rule1} & {Rule2}       & {Rule1} & {Rule2}  & {Rule1} & {Rule2} & {Rule1} & {Rule2} & {Rule1} & {Rule2} & {Rule1} & {Rule2} \\
        \midrule
\multirow{5}{*}{Syn-data1}  & CATE$\uparrow$ & 10.00\tiny{(0.65)} & 8.39\tiny{(0.25)}  & 8.97\tiny{(0.61)} & 7.36\tiny{(0.88)} & 7.42\tiny{(0.63)}  & 6.33\tiny{(1.45)} & 7.27\tiny{(0.31)}  & 6.97\tiny{(0.37)} & 6.93\tiny{(0.24)}  & 6.72\tiny{(0.25)} & 7.53\tiny{(0.55)}  & 7.13\tiny{(0.09)} & 8.16\tiny{(0.25)}  & 7.60\tiny{(0.29)} \\

  & Avg\_ITE$\uparrow$ & 9.17\tiny{(0.38)} & 7.98\tiny{(0.69)} & 7.79\tiny{(0.52)} & 5.89\tiny{(0.72)}& 7.69\tiny{(0.68)} & 6.15\tiny{(1.08)} & 6.81\tiny{(0.50)} & 6.95\tiny{(0.96)} & 6.75\tiny{(0.44)} & 7.22\tiny{(0.52)} & 7.57\tiny{(0.45)} & 7.17\tiny{(0.27)} & 7.99\tiny{(0.13)} & 7.82\tiny{(0.62)}\\

  & Variance$\downarrow$ & 9.20\tiny{(2.24)} & 9.69\tiny{(3.08)} & 10.32\tiny{(2.60)} & 10.37\tiny{(1.55)} & 8.94\tiny{(1.36)} & 7.92\tiny{(2.18)} & 9.79\tiny{(2.90)} & 9.85\tiny{(1.73)} & 8.32\tiny{(3.14)} & 9.23\tiny{(0.94)} & 11.59\tiny{(2.30)} & 12.07\tiny{(1.70)} & 11.60\tiny{(2.21)} & 9.91\tiny{(2.15)}\\

  & PEHE$\downarrow$ & 2.25\tiny{(0.39)} & 2.38\tiny{(0.33)} & 2.46\tiny{(0.14)} & 2.62\tiny{(0.05)} & 2.17\tiny{(0.15)} & 1.84\tiny{(0.20)} & 2.14\tiny{(0.28)} & 2.23\tiny{(0.16)} & 1.94\tiny{(0.24)} & 2.13\tiny{(0.17)} & 2.22\tiny{(0.06)} & 2.36\tiny{(0.10)} & 2.28\tiny{(0.20)} & 2.23\tiny{(0.11)}\\
  
  & MAPE$\downarrow$ & 0.23\tiny{(0.05)} & 0.28\tiny{(0.06)} & 0.33\tiny{(0.03)} & 0.56\tiny{(0.10)}& 0.23\tiny{(0.06)} & 0.28\tiny{(0.05)} & 0.32\tiny{(0.08)} & 0.30\tiny{(0.07)} & 0.27\tiny{(0.03)} & 0.24\tiny{(0.04)}  & 0.27\tiny{(0.02)} & 0.31\tiny{(0.03)} & 0.26\tiny{(0.03)} & 0.25\tiny{(0.05)}\\

\hline
\multirow{5}{*}{Syn-data2} & CATE$\uparrow$  & 12.72\tiny{(0.55)} & 11.44\tiny{(0.69)} & 10.95\tiny{(0.50)} & 9.93\tiny{(0.48)} & 11.30\tiny{(0.58)} & 10.89\tiny{(0.62)} & 10.23\tiny{(0.44)} & 9.89\tiny{(0.29)} & 9.40\tiny{(0.46)} & 8.68\tiny{(0.37)} & 10.24\tiny{(0.07)} & 10.05\tiny{(0.06)} & 11.55\tiny{(0.25)} & 10.99\tiny{(0.50)} \\

  & Avg\_ITE$\uparrow$ & 11.23\tiny{(0.64)} & 10.53\tiny{(0.75)} & 8.99\tiny{(0.49)} & 8.55\tiny{(0.52)} & 10.29\tiny{(0.42)} & 9.70\tiny{(0.69)} & 9.89\tiny{(0.52)} & 10.11\tiny{(1.27)} & 9.39\tiny{(0.61)} & 8.87\tiny{(0.69)} & 9.60\tiny{(0.30)} & 9.50\tiny{(0.27)} & 10.42\tiny{(0.62)} & 10.06\tiny{(0.61)} \\

  & Variance$\downarrow$ & 11.15\tiny{(2.91)} & 10.65\tiny{(3.61)} & 12.79\tiny{(2.43)} & 14.37\tiny{(1.08)} & 11.78\tiny{(3.76)} & 15.28\tiny{(2.08)} & 13.15\tiny{(2.55)} & 12.51\tiny{(1.27)} & 8.71\tiny{(4.18)} & 11.79\tiny{(3.12)} & 13.78\tiny{(1.85)} & 14.50\tiny{(2.56)} & 11.82\tiny{(2.92)} & 14.53\tiny{(2.39)} \\

  & PEHE$\downarrow$ & 2.64\tiny{(0.49)} & 2.13\tiny{(0.31)} & 2.93\tiny{(0.24)} & 2.62\tiny{(0.29)} & 2.32\tiny{(0.35)} & 2.55\tiny{(0.19)} & 2.19\tiny{(0.24)} & 2.47\tiny{(0.33)} & 2.10\tiny{(0.30)} & 1.95\tiny{(0.13)} & 2.29\tiny{(0.19)} & 2.30\tiny{(0.20)} & 2.46\tiny{(0.39)} & 2.51\tiny{(0.09)} \\

  & MAPE$\downarrow$ & 0.23\tiny{(0.06)} & 0.19\tiny{(0.03)} & 0.34\tiny{(0.05)} &0.32\tiny{(0.06)} & 0.21\tiny{(0.05)} & 0.25\tiny{(0.04)} & 0.20\tiny{(0.04)} & 0.22\tiny{(0.11)} & 0.19\tiny{(0.04)} & 0.19\tiny{(0.04)} & 0.22\tiny{(0.03)} & 0.23\tiny{(0.03)} & 0.22\tiny{(0.06)} & 0.19\tiny{(0.04)} \\

  \hline
\multirow{5}{*}{Syn-data3}  & CATE$\uparrow$ & 14.06\tiny{(0.25)} & 12.70\tiny{(1.55)} & 14.56\tiny{(0.54)} & 13.65\tiny{(0.70)} & 12.19\tiny{(0.85)} & 11.15\tiny{(1.06)} & 13.37\tiny{(0.46)} & 12.73\tiny{(0.60)} & 11.90\tiny{(0.73)} & 11.03\tiny{(0.67)} & 12.73\tiny{(0.11)} & 12.56\tiny{(0.03)} & 13.53\tiny{(0.26)} & 13.22\tiny{(0.15)}\\

  & Avg\_ITE$\uparrow$ & 13.80\tiny{(0.61)} & 12.47\tiny{(1.91)} & 12.87\tiny{(0.16)} &  12.08\tiny{(0.97)}& 12.74\tiny{(1.86)} & 11.08\tiny{(1.34)} & 12.73\tiny{(1.72)} & 11.89\tiny{(1.47)} & 12.53\tiny{(0.68)} & 11.75\tiny{(1.15)} & 12.74\tiny{(0.22)} & 12.91\tiny{(0.33)} & 13.31\tiny{(0.32)} & 13.90\tiny{(0.47)}  \\
  
  & Variance$\downarrow$ & 16.18\tiny{(4.44)} & 16.36\tiny{(5.34)} & 19.31\tiny{(4.81)} & 19.10\tiny{(3.38)}& 23.45\tiny{(9.63)} & 16.42\tiny{(4.48)} & 22.36\tiny{(7.74)} & 19.99\tiny{(3.78)} & 18.06\tiny{(2.88)} & 17.48\tiny{(1.87)} & 20.83\tiny{(1.43)} & 21.92\tiny{(3.52)} & 20.04\tiny{(2.33)} & 23.58\tiny{(7.62)}\\

  & PEHE$\downarrow$ & 2.99\tiny{(0.19)} & 3.08\tiny{(0.38)} & 3.38\tiny{(0.27)} & 3.37\tiny{(0.12)} & 3.76\tiny{(0.70)} & 2.75\tiny{(0.23)} & 3.52\tiny{(0.84)} & 3.30\tiny{(0.32)} & 3.17\tiny{(0.22)} & 2.92\tiny{(0.40)} & 3.10\tiny{(0.21)} & 3.24\tiny{(0.22)} & 3.09\tiny{(0.34)} & 3.39\tiny{(0.50)}\\

  & MAPE$\downarrow$ & 0.19\tiny{(0.02)} & 0.23\tiny{(0.07)} & 0.26\tiny{(0.03)} & 0.28\tiny{(0.04)} & 0.25\tiny{(0.03)} & 0.23\tiny{(0.03)} & 0.26\tiny{(0.12)} & 0.28\tiny{(0.08)} & 0.21\tiny{(0.04)} & 0.21\tiny{(0.03)} & 0.21\tiny{(0.02)} & 0.21\tiny{(0.02)} & 0.20\tiny{(0.01)} & 0.20\tiny{(0.01)}\\

  \hline
\multirow{5}{*}{IHDP} & CATE$\uparrow$  & 10.98\tiny{(1.72)} & 9.98\tiny{(0.45)} & 6.52\tiny{(0.37)} & 3.19\tiny{(0.87)} & 8.11\tiny{(0.47)} & 7.75\tiny{(0.61)} & / & / & 8.79\tiny{(1.33)} & 6.90\tiny{(1.14)} & 4.24\tiny{(0.39)} & 4.05\tiny{(0.26)} & 3.36\tiny{(0.80)} & 2.06\tiny{(nan)} \\

  & Avg\_ITE$\uparrow$ & 8.44\tiny{(0.70)} & 8.51\tiny{(0.71)} & 6.08\tiny{(0.36)} & 3.15\tiny{(1.26)} & 6.39\tiny{(1.10)} & 7.01\tiny{(0.57)} & / & / & 7.84\tiny{(0.65)} & 6.22\tiny{(0.55)} & 4.39\tiny{(0.40)} & 4.19\tiny{(0.44)} & 3.47\tiny{(0.60)} & -0.93\tiny{(nan)} \\

  & Variance$\downarrow$ & 4.17\tiny{(5.03)} & 24.06\tiny{(16.35)} & 139.72\tiny{(31.36)} & 210.22\tiny{(33.57)} & 19.36\tiny{(29.37)} & 148.97\tiny{(130.44)} & / & / & 28.80\tiny{(59.60)} & 126.27\tiny{(104.86)} & 172.14\tiny{(17.54)} & 166.78\tiny{(28.04)} & 173.12\tiny{(28.99)} & 198.46\tiny{(nan)} \\

  & PEHE$\downarrow$ & 10.78\tiny{(1.22)} & 10.42\tiny{(1.57)} & 8.58\tiny{(0.39)} & 10.12\tiny{(2.78)} & 8.95\tiny{(1.63)} & 8.59\tiny{(1.15)} & / & / & 9.96\tiny{(1.75)} & 7.36\tiny{(1.64)} & 9.98\tiny{(1.51)} & 9.75\tiny{(1.66)} & 9.34\tiny{(0.88)} & 23.10\tiny{(nan)} \\

  & MAPE$\downarrow$ & 2.03\tiny{(0.65)} & 2.08\tiny{(0.94)} & 1.86\tiny{(0.34)} & 1.25\tiny{(0.29)} & 1.58\tiny{(0.21)} & 1.54\tiny{(0.34)} & / & / & 2.01\tiny{(0.90)} & 2.21\tiny{(1.74)} & 1.62\tiny{(0.41)} & 1.51\tiny{(0.37)} & 1.32\tiny{(0.26)} & 2.29\tiny{(nan)} \\

        \bottomrule
    \end{tabular}
    \end{adjustbox}
    \label{tbl:rules}
    \vspace{-2mm}
\end{table*}

\textbf{Metrics}. In order to evaluate the effectiveness of the causal rules. On the one hand, we assess the subgroup treatment effect from the perspectives of effect strength, uncertainty, and accuracy. The specific metrics are described as follows: (1) Estimated CATE; (2) True CATE (mean value of ITE within subgroup); (3) The variance of the outcome of the treated units in the subgroup. (4) The precision in the estimation of heterogeneous effects $\text{PEHE}=\sqrt{\frac{1}{n}\sum_{i=1}^{n}(\hat\tau(\textbf{x}_i) - \tau(\textbf{x}_i))^2}$; (5) The mean absolute percentage error $\text{MAPE}=\frac{1}{n}\sum_{i=1}^n|\frac{\hat\tau(\textbf{x}_i) - \tau(\textbf{x}_i)}{\tau(\textbf{x}_i)}|$. On the other hand, we have also measured the interpretability of the rule set, including the following metrics: (1) Average length of rule antecedent; (2) Average overlap between pairs of rules; (3) Rule set coverage.

\textbf{Implement detail}. We used 5-fold cross-validation and Bayesian optimization to tune parameters. Specifically, we optimize the parameters of \algname with max rule length $L \in \{3,4,5,6\}$, variance weight $\lambda \in \{0.1, \cdots, 1.5\}$. CT’s hyperparameters include cross-validation method cv.option=``matching'' and pruning factor pru\_coef $\in \{0.4, 0.9, 1.5\}$. For CF, its hyperparameter takes the values num.trees $\in \{5, 8, 10\}$, honest version of the CT split.Honest=TRUE and tradeoff between effect and variance split.alpha $\in \{0.2, 0.5, 0.8\}$. The CRE parameters include ntrees $\in \{20, 25\}$, max\_depth $\in \{3, 4\}$ and the decay threshold for rules pruning t\_decay $\in \{0.025, 0.01, 0.04\}$. For DT, the depth of tree max\_depth is fixed as 4. In PYS, the result set has 10 rules, with a maximum rule depth of $\{2, 5\}$, using the subgroup scoring method qf=ps.WRAccQF(). For BRCG, we tune the maximum number of columns generated per iteration K from 8 to 12, and the max rule length is chosen from $\{5, 10\}$.

\textbf{Results}. The evaluation results are reported in Table~\ref{tbl:rules}. We recorded the estimated CATE and the ground truth CATE (Avg\_ITE) to assess the strength of the treatment effect. The results show that the estimated CATE and true CATE of both rules of \algname is about 16.1\% and 13.8\% higher than the effect values of other baselines. We also observe that for correlation learning methods, such as DT and PYS, the highest effect is not the rule with a high predicted value of $Y$ (in our experiment, it is the rules with the probability of $Y$ around 0.6-0.8), which reflects the difference between correlation and causation. When it comes to variance, \algname reduces variance by about 12.0\% compared to other baselines. In addition, \algname sacrifices certain effects to reduce variance when necessary. For example, in rule2 of data3, the effect and variance of \algname are 12.47 and 16.36, while the effect of PYS and BRCG are 12.91 and 13.90, respectively, which are larger than \algname, but their variance is also large at 21.92 and 23.58. We also compared PEHE and MAPE, which measure the accuracy of ITE estimates. The results show that the estimation accuracy of \algname is, on average, 0.05\% higher on PEHE and 1.6\% smaller on MAPE compared to the other methods. This suggests that \algname is able to find subgroups with more significant treatment effects with similar or better estimation accuracy. It is worth noting that while CATE is the best estimate of ITE in terms of the mean squared error~\cite{kunzel2019metalearners}, it can also lead to inaccurate estimates because the estimation method we used, IPW, inherently has errors in estimating when the propensity score approaches 0 or 1. A potential solution is to introduce more robust estimators, such as doubly robust estimation~\cite{Funk2011DoublyRE}.

Table~\ref{tbl:ruleset} shows the relevant metrics on the rule set readability. We found that the average rule length of \algname is around 3, which is mostly smaller than tree-based methods such as CT and CF. In addition, the overlap between rules in \algname is also at a low level of 0.1\%, which is favorable for user understanding. The coverage metric shows that \algname focuses on a small number of groups with strong effects, while other methods, like CT and CRE, have coverage rates as high as more than 50\%. However, the variance of their coverage is also very large, indicating that there are significant differences between their rules. Also, the high coverage may be responsible for their more average treatment effects.

\begin{table}[htb]
    \centering
    \caption{Interpretability metrics for the rule set, reported as mean and standard deviation.}
    % \small
    \vspace{-2mm}
    \setlength{\tabcolsep}{2.7pt}
    % \begin{adjustbox}{width=\textwidth}
        \begin{tabular}{
            l
            |l
            |S[table-format=1.1, input-symbols=()]
            |S[table-format=1.1, input-symbols=()]
            |S[table-format=1.1, input-symbols=()]
            |S[table-format=1.1, input-symbols=()]
            |S[table-format=1.1, input-symbols=()]
            |S[table-format=1.1, input-symbols=()]
            |S[table-format=1.1, input-symbols=()]
        }
        \toprule
        {Dataset} & {Metrics} & {CURLS} & { CT} & { CF} & {CRE} & { DT} & {PYS} & {BRCG} \\
        \midrule
\multirow{6}{*}{Syn-data1}  & \multirow{2}{*}{Avg\_len} & 2.9 & 2.3 & 5.1 & 2.1 & 4.0 & 1.1 & 2.3 \\ [-0.5em] 
       & & \tiny{(0.8)} & \tiny{(0.8)} & \tiny{(1.5)} & \tiny{(0.7)} & \tiny{(0.0)} & \tiny{(0.2)} & \tiny{(0.3)} \\
\cline{2-9}
  & \multirow{2}{*}{Overlap(\%)} & 0.1 & 0.0 & 0.0 & 1.4 & 0.0 & 3.7 & 1.0\\ [-0.5em] 
       & & \tiny{(0.1)} & \tiny{(0.0)} & \tiny{(0.1)} & \tiny{(0.7)} & \tiny{(0.0)} & \tiny{(0.7)} & \tiny{(0.9)} \\
\cline{2-9}
  & \multirow{2}{*}{Coverage(\%)} & 6.0 & 46.1 & 10.6 & 24.1 & 14.5 & 31.3 & 16.5 \\ [-0.5em] 
      &  & \tiny{(0.6)} & \tiny{(41.5)} & \tiny{(2.8)} & \tiny{(6.3)} & \tiny{(1.4)} & \tiny{(6.5)} & \tiny{(6.7)} \\

\hline

\multirow{6}{*}{Syn-data2} & \multirow{2}{*}{Avg\_len} & 3.0 & 5.9 & 2.7 & 1.9 & 4.0 & 1.0 & 2.5 \\ [-0.5em] 
      & & \tiny{(0.0)} & \tiny{(2.1)} & \tiny{(0.8)} & \tiny{(0.5)} & \tiny{(0.0)} & \tiny{(0.0)} & \tiny{(0.4)}\\
\cline{2-9}
    & \multirow{2}{*}{Overlap(\%)} & 0.2 & 0.0 & 1.2 & 1.1 & 0.0 & 4.1 & 2.1 \\ [-0.5em] 
    & & \tiny{(0.2)} & \tiny{(0.0)} & \tiny{(0.8)} & \tiny{(2.2)} & \tiny{(0.0)} & \tiny{(1.0)} & \tiny{(1.8)} \\
\cline{2-9}
    & \multirow{2}{*}{Coverage(\%)} & 7.4 & 36.9 & 11.7 & 38.9 & 14.7 & 36.4 & 13.9 \\ [-0.5em] 
    & & \tiny{(1.0)} & \tiny{(22.2)} & \tiny{(1.6)} & \tiny{(34.3)} & \tiny{(1.9)} & \tiny{(2.2)} & \tiny{(1.4)} \\

\hline

\multirow{6}{*}{Syn-data3}  & \multirow{2}{*}{Avg\_len} & 2.8 & 4.6 & 5.3 & 1.4 & 4.0 & 1.0 & 2.3 \\ [-0.5em] 
      &  & \tiny{(0.5)} & \tiny{(2.0)} & \tiny{(1.2)} & \tiny{(0.6)} & \tiny{(0.0)} & \tiny{(0.0)} & \tiny{(0.7)}  \\
\cline{2-9}
  & \multirow{2}{*}{Overlap(\%)} & 0.1 & 0.0 & 0.5 & 0.5 & 0.0 & 3.7 & 0.6 \\ [-0.5em] 
      &  & \tiny{(0.2)} & \tiny{(0.0)} & \tiny{(0.6)} & \tiny{(1.2)} & \tiny{(0.0)} & \tiny{(0.6)} & \tiny{(0.1)} \\
\cline{2-9}
  & \multirow{2}{*}{Coverage(\%)} & 12.1 & 27.5 & 8.4 & 68.1 & 16.4 & 35.8 & 14.7\\ [-0.5em] \textbf{}
      &  & \tiny{(1.4)} & \tiny{(21.4)} & \tiny{(1.6)} & \tiny{(36.5)} & \tiny{(3.8)} & \tiny{(2.7)} & \tiny{(1.1)} \\

\hline

\multirow{6}{*}{IHDP}  & \multirow{2}{*}{Avg\_len} & 3.0 & 4.7 & 3.9 & / & 4.0 & 4.8 & 4.3 \\ [-0.5em] 
      &  & \tiny{(0.8)} & \tiny{(2.3)} & \tiny{(1.6)} &  & \tiny{(0.0)} & \tiny{(0.3)} & \tiny{(0.8)}  \\
\cline{2-9}
  & \multirow{2}{*}{Overlap(\%)} & 0.7 & 0.0 & 5.0 & / & 0.0 & 41.6 & 0.0 \\ [-0.5em]
       &  & \tiny{(0.8)} & \tiny{(0.0)} & \tiny{(6.8)} &  & \tiny{(0.0)} & \tiny{(9.5)} & \tiny{(0.0)}  \\
\cline{2-9}
  & \multirow{2}{*}{Coverage(\%)} & 8.3 & 25.8 & 11.5 & / & 21.3 & 55.8 & 53.5 \\ [-0.5em] 
       &  & \tiny{(1.8)} & \tiny{(24.6)} & \tiny{(3.4)} &  & \tiny{(10.9)} & \tiny{(5.0)} & \tiny{(8.8)}  \\
    
        \bottomrule
    \end{tabular}
    \label{tbl:ruleset}
    \vspace{-2mm}
\end{table}

\subsection{Case Studies}

We qualitatively evaluate the performance of \algname on two real and easily understandable accident analysis and policy making datasets. The Titanic dataset\footnote{\url{https://www.kaggle.com/c/titanic/data}} provides passenger data on survival, sex, age, fares, number of siblings/spouses on board (sibsp) and number of parents/children (parch) \etc We want to determine how premium class (treatment) affects passenger survival (outcome). For the treatment, $T=1$ means that the passenger is in premium class 1 and 2 cabins, and $T=0$ means the lowest class 3 cabin. We extracted three causal rules from the dataset; the first shows that upgrading to a higher class improves survivability for passengers with more family members, who may be more willing to aid each other. The second and third rules correspond to the fact that women and children who pay higher ticket fees are more likely to be rescued due to the abundant rescue resources in the higher class and the ``women and children first'' policy.

The Lalonde dataset (part of the famous Jobs dataset)~\cite{Lalonde1984EvaluatingTE} includes data from participants and non-participants in a job training program (National Supported Work Demonstration, NSW). NSW is an experimental program that aims to help economically disadvantaged people (\eg long-term unemployed, high school dropouts) return to work. It would train underprivileged workers in work skills for 9-18 months. We evaluated how the training program (treatment) affected income (outcome). The covariates include individual background information (age, race, academic background, and past income). Table~\ref{tbl:cases} reveals two subgroups with high treatment effects yielded by \algname. The first subgroup is married people over 29 who may be living a stressful life and will study hard to increase their income during training. The second subgroup is 18-19-years-olds. They have good learning capacity and ambition, so they can get high-paying employment through training despite their lack of experience. We used DoWhy\footnote{\url{https://github.com/py-why/dowhy}}, a famous causal inference package, to calculate the treatment effect for the entire population, which is 1639.8, less than half of the two mentioned subgroups. With reference to causal rules, policymakers can better choose target groups to improve program implementation outcomes.

\begin{table}[htb]
        \centering
        \caption{Examples of learned causal rules.}
        \vspace{-2mm}
        % \begin{adjustbox}{width=\linewidth}
        \begin{tabular}{c}
        \toprule
        Titanic \\
        \midrule
        IF sibsp > 1.0 AND parch > 1.0 THEN $\tau$ = 0.88 \\
        IF sex == female AND fare > 39.7 THEN $\tau$ = 0.81 \\
        IF age <= 20.0 AND fare > 21.7 THEN $\tau$ = 0.75 \\
        \hline\hline
        Lalonde \\
        \midrule
        IF age > 29.0 AND married == 1.0 THEN $\tau$ = 4722.3 \\
        IF age <= 19.0 AND age > 18.0 THEN $\tau$ = 4165.8 \\
        \bottomrule
        \end{tabular}
        % \end{adjustbox}
        \label{tbl:cases}
        \vspace{-2mm}
\end{table}

%% file: sections/7_discussion.tex
\section{Discussion}

In this section, we discuss the implications, scalability, limitations and future work of \algname.

\textbf{Implications.} This research leads to two key implications. Firstly, causal rule learning is helpful, and exploration of its algorithmic design is encouraging. The concise rules in the form of "IF-THEN" are similar to the logic of human decision-making. They are easier to understand compared to complicated tree structures and black-box methods for treatment effect estimation, making it simpler to discover new causal knowledge. Secondly, \algname can be adopted in practical applications, as our real-world data cases demonstrate. In addition to those examples, \algname can also be adapted to scenarios requiring causal-assisted decision-making, such as education and industry. For example, \algname may help teachers find the best way to teach different students based on their characteristics. It may also assist quality inspectors troubleshoot metric anomalies and attribute them to specific models.

\begin{figure}[htb]
\vspace{-2mm}
  \centering
  \includegraphics[width=\linewidth]{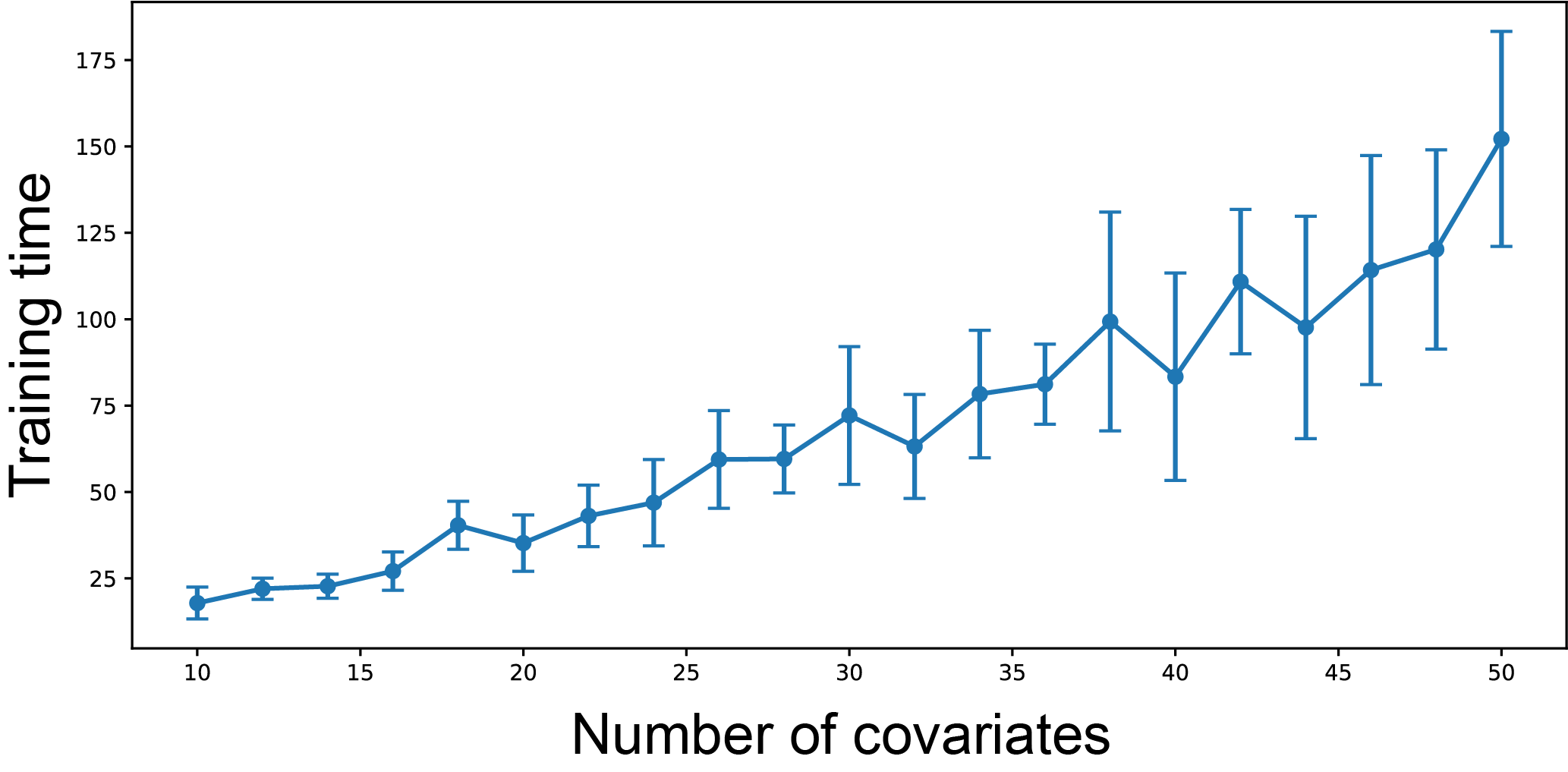}
  \caption{Scalability test. Training time scales linearly with the number of covariates.}
  
  \label{fig:scalability}
  \vspace{-2mm}
\end{figure}

\textbf{Scalability.} The computational complexity of \algname is mainly dominated by local search in Alg.~\ref{alg3}, which is linear in the number of covariates. To demonstrate this empirically, we conducted an experiment to measure the algorithm running time under different numbers of covariates and units. As shown in Fig.~\ref{fig:scalability}, the training time grows linearly with the number of covariates. 
% Therefore, \algname is scalable with respect to the amount of data and the number of covariates.

\textbf{Limitations and future work.} We believe there are three potential directions that \algname can further explore. First, \algname may converge to suboptimal results due to its iterative optimization procedure, where poor initialization and iteration paths can lead to local optima. To address this, we use a greedy strategy for initialization for MM and incorporate local search techniques at the end. We are also exploring other optimization methods, such as neural combinatorial optimization~\cite{Bello2016NeuralCO} and multi-objective learning~\cite{9527391}, to further improve the quality of the final solution. Second, the descriptive ability of antecedents is limited. While \algname uses a form of CNF with AND logical connectives and binary covariates, its inability to handle OR connectives and limitations in discretization may restrict it from describing certain refined subgroups. Future work involves integrating logical connectives into the learning process and adaptively determining the discretization. Finally, the assumptions on single treatment and single outcome may not be compatible with real-world scenarios. In practice, there may be multiple treatments, unobserved variables, multiple outcomes, or more complex causal relationships. For non-binary treatments, we can extend our method by utilizing One-Versus-The-Rest (OvR), which is commonly used for multi-classification tasks, to handle each treatment value. Other methods such as robust HTE~\cite{Kennedy2020TowardsOD} and instrumental variables~\cite{syrgkanis2019machine} can also be investigated to ensure the validity of causal inference.

%% file: sections/8_conclusion.tex
\section{Conclusion}

In this paper, we propose a new method called \algname for learning causal rules from observational data. To the best of our knowledge,  this is the first method that employs an optimization problem to generate rules to explain subgroups with significant treatment effects. We formally define these causal rules composed of antecedents that form conditions to characterize the subgroup and the associated effects. We model the rule learning process as a discrete optimization problem. By constructing an approximate submodular lower bound for the original objective, the problem can be solved iteratively based on the minorize-maximization algorithm. Quantitative experiments and qualitative case studies demonstrate that our method is effective in identifying meaningful causal rules from observational data. Future works involve more effective optimization algorithms, refining rule formation, and addressing more complex scenarios.

%% file: sections/9_appendix.tex
\section{Supplemental Experiments}

The above experiments compares \algname to popular tree-based CATE methods (CT, CF), causal rule learning method (CRE), subgroup discovery methods (DT, PYS), and rule learning method (BRCG). The path from the root to the leaf nodes can be viewed as a description of the subgroups in these tree- or rule-based methods, whereas other black-box causal heterogeneity or uplift modeling models usually lack interpretability and are not used for the comparison. As a workaround, we supplemented new baselines, including double machine learning, doubly robust, and orthogonal random forest. We train a decision tree on the effects of these estimators using Tree Interpreter from a popular causal inference package, EconML~\cite{econml}, to indirectly obtain subgroup descriptions. New baselines includes:
\begin{itemize}
    % \item IHDP: This dataset is a semi-synthetic dataset constructed from the Infant Health and Development Program (IHDP).  It is a randomized experiment from 1985 to 1988 which studied the effect of home visits on cognitive test scores for infants. Our version of dataset is the dataset used by Louizos \etal~\cite{louizos2017causal}. This is the first realization of 10 generated datasets and you can find other realizations from \url{https://github.com/AMLab-Amsterdam/CEVAE}. It contains 7470 units, 19 binary covariates and 6 continuous covariates.

    \item LDML: The double machine learning estimator with a low-dimensional linear final stage implemented as a statsmodel regression.

    \item DRL: CATE estimator that uses doubly-robust correction techniques to account for covariate shift (selection bias) between the treatment arms.

    \item DROF: Orthogonal random forest for discrete treatments using the doubly robust moment function.
\end{itemize}

As shown in Table \ref{tbl:rules-sup}, \algname still achieves competitive performance, being able to identify subgroups described by rules with stronger treatment effects and smaller outcome variance, while maintaining similar PEHE and MAPE accuracies.

\begin{table*}[htb!]
    \centering
    \caption{5-fold average performance metrics of the new baselines. Numbers in parentheses represent standard deviations.}
    \label{tbl:rules-sup}
    % \begin{adjustbox}{width=\textwidth}
        \begin{tabular}{
            l
            |l
            |S[table-format=2.2,table-space-text-post=\tiny{(3.30)}]
            S[table-format=2.2,table-space-text-post=\tiny{(3.30)}]
            |S[table-format=2.2,table-space-text-post=\tiny{(3.30)}]
            S[table-format=2.2,table-space-text-post=\tiny{(3.30)}]
            |S[table-format=2.2,table-space-text-post=\tiny{(3.30)}]
            S[table-format=2.2,table-space-text-post=\tiny{(3.30)}]
            |S[table-format=2.2,table-space-text-post=\tiny{(3.30)}]
            S[table-format=2.2,table-space-text-post=\tiny{(3.30)}]
        }
        \toprule
        \multirow{2}{*}{Dataset} & \multirow{2}{*}{Metrics} & \multicolumn{2}{c|}{CURLS} & \multicolumn{2}{c|}{LDML} &
        \multicolumn{2}{c|}{DRL} & \multicolumn{2}{c}{DROF} \\
                                 &                          & {Rule1} & {Rule2}         & {Rule1} & {Rule2}       & {Rule1} & {Rule2}  & {Rule1} & {Rule2}  \\
        \midrule
\multirow{5}{*}{Syn-data1}  & CATE$\uparrow$ & 10.00\tiny{(0.65)} & 8.39\tiny{(0.25)} & 8.06\tiny{(0.27)} & 7.28\tiny{(0.15)} & 8.17\tiny{(0.20)} & 7.39\tiny{(0.37)} & 9.51\tiny{(0.47)} & 8.55\tiny{(0.21)} \\

  & Avg\_ITE$\uparrow$ & 9.17\tiny{(0.38)} & 7.98\tiny{(0.69)} & 8.41\tiny{(0.19)} & 7.08\tiny{(0.48)} & 8.42\tiny{(0.44)} & 7.74\tiny{(0.44)} & 8.19\tiny{(0.53)} & 7.58\tiny{(0.57)} \\

  & Variance$\downarrow$ & 9.20\tiny{(2.24)} & 9.69\tiny{(3.08)} & 9.63\tiny{(1.50)} & 10.08\tiny{(3.08)} & 9.10\tiny{(3.33)} & 9.11\tiny{(1.78)} & 9.48\tiny{(4.16)} & 9.89\tiny{(2.01)} \\

  & PEHE$\downarrow$ & 2.25\tiny{(0.39)} & 2.38\tiny{(0.33)} & 2.14\tiny{(0.21)} & 2.15\tiny{(0.23)} & 2.24\tiny{(0.31)} & 1.98\tiny{(0.38)} & 2.69\tiny{(0.15)} & 2.27\tiny{(0.22)} \\
  
  & MAPE$\downarrow$ & 0.23\tiny{(0.05)} & 0.28\tiny{(0.06)} & 0.21\tiny{(0.02)} & 0.29\tiny{(0.06)} & 0.23\tiny{(0.02)} & 0.22\tiny{(0.05)} & 0.35\tiny{(0.06)} & 0.30\tiny{(0.05)} \\

\hline
\multirow{5}{*}{Syn-data2} & CATE$\uparrow$  & 12.72\tiny{(0.55)} & 11.44\tiny{(0.69)} & 10.44\tiny{(0.23)} & 9.54\tiny{(0.20)} & 10.85\tiny{(0.65)} & 10.00\tiny{(0.40)} & 10.80\tiny{(0.45)} & 9.48\tiny{(0.40)} \\

  & Avg\_ITE$\uparrow$ & 11.23\tiny{(0.64)} & 10.53\tiny{(0.75)} & 10.19\tiny{(0.56)} & 9.52\tiny{(0.41)} & 10.66\tiny{(0.54)} & 9.86\tiny{(0.47)} & 9.47\tiny{(0.45)} & 8.91\tiny{(0.53)} \\

  & Variance$\downarrow$ & 11.15\tiny{(2.91)} & 10.65\tiny{(3.61)} & 13.28\tiny{(3.12)} & 13.03\tiny{(1.97)} & 13.51\tiny{(1.93)} & 10.07\tiny{(1.27)} & 14.64\tiny{(4.17)} & 13.73\tiny{(1.38)} \\

  & PEHE$\downarrow$ & 2.64\tiny{(0.49)} & 2.13\tiny{(0.31)} & 2.23\tiny{(0.18)} & 2.20\tiny{(0.25)} & 2.15\tiny{(0.25)} & 1.97\tiny{(0.32)} & 2.67\tiny{(0.30)} & 2.26\tiny{(0.13)} \\

  & MAPE$\downarrow$ & 0.23\tiny{(0.06)} & 0.19\tiny{(0.03)} & 0.19\tiny{(0.03)} & 0.21\tiny{(0.03)} & 0.18\tiny{(0.02)} & 0.17\tiny{(0.04)} & 0.28\tiny{(0.06)} & 0.25\tiny{(0.04)} \\

  \hline
\multirow{5}{*}{Syn-data3}  & CATE$\uparrow$ & 14.06\tiny{(0.25)} & 12.70\tiny{(1.55)} & 14.39\tiny{(0.45)} & 13.30\tiny{(0.07)} & 13.08\tiny{(0.99)} & 12.37\tiny{(0.78)} & 16.02\tiny{(0.69)} & 14.61\tiny{(0.46)} \\

  & Avg\_ITE$\uparrow$ & 13.80\tiny{(0.61)} & 12.47\tiny{(1.91)} & 13.76\tiny{(0.42)} & 12.66\tiny{(0.27)} & 13.88\tiny{(0.46)} & 13.21\tiny{(1.45)} & 13.98\tiny{(0.62)} & 12.96\tiny{(0.45)} \\
  
  & Variance$\downarrow$ & 16.18\tiny{(4.44)} & 16.36\tiny{(5.34)} & 24.67\tiny{(15.93)} & 24.54\tiny{(6.54)} & 23.22\tiny{(3.99)} & 18.32\tiny{(2.76)} & 16.45\tiny{(5.79)} & 20.25\tiny{(3.21)} \\

  & PEHE$\downarrow$ & 2.99\tiny{(0.19)} & 3.08\tiny{(0.38)} & 3.27\tiny{(0.48)} & 3.21\tiny{(0.29)} & 3.13\tiny{(0.34)} & 3.13\tiny{(0.62)} & 3.57\tiny{(0.50)} & 3.35\tiny{(0.40)} \\

  & MAPE$\downarrow$ & 0.19\tiny{(0.02)} & 0.23\tiny{(0.07)} & 0.22\tiny{(0.03)} & 0.24\tiny{(0.02)} & 0.18\tiny{(0.03)} & 0.20\tiny{(0.02)} & 0.25\tiny{(0.05)} & 0.25\tiny{(0.04)} \\

  \hline
\multirow{5}{*}{IHDP}  & CATE$\uparrow$ & 10.98\tiny{(1.72)} & 9.98\tiny{(0.45)} & 2.15\tiny{(0.98)} & 1.29\tiny{(0.82)} & 7.31\tiny{(0.80)} & 6.52\tiny{(0.74)} & 7.70\tiny{(0.45)} & 6.61\tiny{(0.40)} \\

  & Avg\_ITE$\uparrow$ & 8.44\tiny{(0.70)} & 8.51\tiny{(0.71)} & 5.36\tiny{(0.69)} & 3.00\tiny{(2.98)} & 7.74\tiny{(0.82)} & 7.04\tiny{(0.61)} & 8.32\tiny{(1.21)} & 7.84\tiny{(1.23)}  \\
  
  & Variance$\downarrow$ & 4.17\tiny{(5.03)} & 24.06\tiny{(16.35)} & 128.78\tiny{(44.79)} & 153.16\tiny{(51.99)} & 95.34\tiny{(92.46)} & 148.74\tiny{(89.66)} & 79.55\tiny{(53.33)} & 172.09\tiny{(85.54)} \\

  & PEHE$\downarrow$ & 10.78\tiny{(1.22)} & 10.42\tiny{(1.57)} & 8.56\tiny{(2.32)} & 12.49\tiny{(6.05)} & 9.59\tiny{(1.08)} & 9.37\tiny{(0.87)} & 10.24\tiny{(1.61)} & 9.51\tiny{(1.89)} \\

  & MAPE$\downarrow$ & 2.03\tiny{(0.65)} & 2.08\tiny{(0.94)} & 0.75\tiny{(0.17)} & 1.07\tiny{(0.34)} & 1.75\tiny{(0.70)} & 2.18\tiny{(1.27)} & 1.68\tiny{(0.66)} & 1.19\tiny{(0.37)} \\

        \bottomrule
    \end{tabular}
    % \end{adjustbox}
\end{table*}

The evaluation results of the interpretability metrics are shown in Table~\ref{tbl:ruleset-sup}, in which the average length of the rules of \textit{CURLS} is shorter and the overlap rate is small, which helps users to understand. In addition, the coverage rate shows that \textit{CURLS} can find more fine-grained subgroups with significant treatment effects.

\begin{table}[H]
    \centering
    \caption{Results of the interpretability metrics of new baselines, reported as mean and standard deviation.}
    \label{tbl:ruleset-sup}
    % \small
    \setlength{\tabcolsep}{4pt}
    % \begin{adjustbox}{width=\textwidth}
        \begin{tabular}{
            l
            |l
            |S[table-format=1.1, input-symbols=()]
            |S[table-format=1.1, input-symbols=()]
            |S[table-format=1.1, input-symbols=()]
            |S[table-format=1.1, input-symbols=()]
        }
        \toprule
        {Dataset} & {Metrics} & {CURLS} & {LDML} & {DRL} & {DROF} \\
        \midrule
\multirow{6}{*}{Syn-data1}  & \multirow{2}{*}{Avg\_len} & 2.9 & 3.7 & 2.6 & 3.4 \\ [-0.5em] 
       & & \tiny{(0.8)} & \tiny{(0.3)} & \tiny{(0.2)} & \tiny{(0.2)} \\
\cline{2-6}
  & \multirow{2}{*}{Overlap(\%)} & 0.1 & 0.0 & 0.0 & 0.0 \\ [-0.5em] 
       & & \tiny{(0.1)} & \tiny{(0.0)} & \tiny{(0.0)} & \tiny{(0.0)} \\
\cline{2-6}
  & \multirow{2}{*}{Coverage(\%)} & 6.0 & 14.5 & 9.9 & 10.1 \\ [-0.5em] 
      &  & \tiny{(0.6)} & \tiny{(2.7)} & \tiny{(1.5)} & \tiny{(1.8)} \\

\hline

\multirow{6}{*}{Syn-data2} & \multirow{2}{*}{Avg\_len} & 3.0 & 3.7 & 2.5 & 4.0 \\ [-0.5em] 
      & & \tiny{(0.0)} & \tiny{(0.4)} & \tiny{(0.4)} & \tiny{(0.0)} \\
\cline{2-6}
    & \multirow{2}{*}{Overlap(\%)} & 0.2 & 0.0 & 0.0 & 0.0 \\ [-0.5em] 
    & & \tiny{(0.2)} & \tiny{(0.0)} & \tiny{(0.0)} & \tiny{(0.0)}  \\
\cline{2-6}
    & \multirow{2}{*}{Coverage(\%)} & 7.4 & 19.0 & 14.7 & 23.5 \\ [-0.5em] 
    & & \tiny{(1.0)} & \tiny{(4.2)} & \tiny{(3.8)} & \tiny{(8.3)} \\

\hline

\multirow{6}{*}{Syn-data3}  & \multirow{2}{*}{Avg\_len} & 2.8 & 3.6 & 2.9 & 3.4 \\ [-0.5em] 
      &  & \tiny{(0.5)} & \tiny{(0.2)} & \tiny{(0.7)} & \tiny{(0.2)}  \\
\cline{2-6}
  & \multirow{2}{*}{Overlap(\%)} & 0.1 & 0.0 & 0.0 & 0.0 \\ [-0.5em] 
      &  & \tiny{(0.2)} & \tiny{(0.0)} & \tiny{(0.0)} & \tiny{(0.0)} \\
\cline{2-6}
  & \multirow{2}{*}{Coverage(\%)} & 12.1 & 14.6 & 12.4 & 12.1 \\ [-0.5em] 
      &  & \tiny{(1.4)} & \tiny{(3.7)} & \tiny{(1.8)} & \tiny{(2.1)} \\

\hline

\multirow{6}{*}{IHDP}  & \multirow{2}{*}{Avg\_len} & 3.0 & 3.8 & 4.0 & 4.0 \\ [-0.5em] 
      &  & \tiny{(0.8)} & \tiny{(0.4)} & \tiny{(0.0)} & \tiny{(0.0)}  \\
\cline{2-6}
  & \multirow{2}{*}{Overlap(\%)} & 0.7 & 0.0 & 0.0 & 0.0 \\ [-0.5em] 
      &  & \tiny{(0.8)} & \tiny{(0.0)} & \tiny{(0.0)} & \tiny{(0.0)} \\
\cline{2-6}
  & \multirow{2}{*}{Coverage(\%)} & 8.3 & 22.4 & 22.7 & 14.8 \\ [-0.5em] 
      &  & \tiny{(1.8)} & \tiny{(20.8)} & \tiny{(12.1)} & \tiny{(4.1)} \\

        \bottomrule
    \end{tabular}
\end{table}

\section{Theoretical Analysis of Approximate Bounds}

We analyse the approximation bounds for $g(\mathcal{R})$. In Proposition \ref{proposition3}, we have shown that $g(\mathcal{R})$ is submodular. Since the parts of $g(\mathcal{R})$ are either increasing (\eg $Q_{3, \mathcal{R}}$) or decreasing (\eg $-Q_{2, \mathcal{R}}$) with the addition of covariates, $g(\mathcal{R})$ is not necessarily monotone. There is a $\frac{1}{k + 2 + \frac{1}{k} + \epsilon}$-approximation bound for non-monotone submodular functions under $k$ matroid constraints~\cite{10.1145/1536414.1536459}. Our problem formulation uses a cardinality constraints $|\mathcal{R}| \le L$, which can be viewed as a $k=1$ matroid constraint; hence $g(\mathcal{R})$ has $\frac{1}{4}$-approximation bound.